\documentclass{article}
\PassOptionsToPackage{numbers,sort&compress}{natbib}

\IfFileExists{neurips_2026.sty}{
 \usepackage[main]{neurips_2026}
}{
 \usepackage[margin=1in]{geometry}
 \usepackage[numbers,sort&compress]{natbib}
}

\usepackage[utf8]{inputenc}
\usepackage[T1]{fontenc}
\usepackage{hyperref}
\usepackage{url}
\usepackage{xurl}
\usepackage{booktabs}
\usepackage{amsfonts}
\usepackage{nicefrac}
\usepackage{microtype}
\usepackage{xcolor}
\usepackage{amsmath}
\usepackage{amssymb}
\usepackage{graphicx}
\usepackage{subcaption}
\usepackage{multirow}
\usepackage{enumitem}
\usepackage{amsthm}
\usepackage{makecell}
\usepackage{threeparttable}

\DeclareMathOperator{\Score}{Score}
\DeclareMathOperator{\TopOneProj}{Top1Proj}
\DeclareMathOperator{\TopOneCos}{Top1Cos}

\title{Compact-Memory LLM Agents via Online Max-Member Clustering and Atom-Aware Packing}

\author{%
Jiahe Geng, Jinpeng Wang, Kun Yuan \\
}

\begin{document}

\maketitle

\begin{abstract}
Many practical long-horizon LLM deployments face tight prompt budgets because latency, cost, and context limits make full-context prompting impractical as interaction length grows. In that setting, the key question is not raw recall alone, but which memory design gives the best quality--token trade-off in the compact-memory regime. To answer this question, we present \textbf{RSM-full}, an online clustered-memory pipeline designed for a strong quality--token Pareto operating point.

RSM-full combines two simple design choices: a cosine-gated \emph{max-member merge} write rule and an atom-aware grouped context packer. On AMA-Bench, our primary compact-memory benchmark, it reaches $83\%$ of Full-Context quality at $32\%$ of the token cost at a $4$k prompt budget; under matched four-seed averaging it outperforms the closest streaming-clustered baseline (Online K-Means) by $+3.5$--$6.0$\,pp ($p{<}.001$) across the entire ${\sim}2.6$k--${\sim}5$k regime. Matched three-seed ablations show that most of this advantage comes from the merge rule ($+5.7$\,pp over Online K-Means and matched-$\tau$ DP-means) and the grouped packer ($+5.0$\,pp over flat concatenation).

The same compact-memory pattern reproduces on RealMem, an independent long-horizon persona-memory benchmark: RSM-full improves on Budget-RAG ($+0.69$\,pp, $p{=}.006$), is on par with BM25-RAG (paired $\Delta{=}{+}0.27$\,pp, $p{=}.47$, no significant difference; we do \emph{not} claim BM25 equivalence in the equivalence-test sense), and significantly outperforms Streaming-Proto ($+2.97$\,pp) and the closest reproduced 2025 agentic-memory baseline A-MEM ($+1.65$\,pp, $p{<}.001$). Across benchmarks, the main empirical message is consistent: under tight prompt budgets, compact-memory performance is driven mainly by how streaming memories are merged and how retrieved content is assembled.

Overall, RSM-full is most useful when answer quality must be preserved under roughly $2$k--$5$k prompt tokens, where it defines a strong compact-memory Pareto point. Higher-token baselines remain stronger outside this regime.
\end{abstract}

\section{Introduction}
Many practical long-horizon LLM deployments must retain past interaction while meeting latency, cost, and context-window constraints. Full-context prompting preserves answer quality, but its token cost grows linearly with interaction length. Fixed-window or top-$K$ retrieval controls prompt length, but often discards the latent structure that makes prior experience reusable~\citep{lewis2020rag,gao2023retrieval,yao2023react}. The central question is therefore not raw recall in isolation, but \emph{which memory operating point yields the best quality--token trade-off in the compact-memory regime ($\le 5$k prompt tokens)?}

We answer this question with \textbf{RSM-full}, an online clustered-memory pipeline built around two simple design choices: a cosine-threshold \emph{max-member merge} write rule and an atom-aware grouped context packer. The paper's main claim is that these choices define a strong compact-memory Pareto point under tight prompt budgets. The empirical spine is two positive end-to-end benchmarks: AMA-Bench as the primary compact-memory benchmark, and RealMem as an independent replication on long-horizon persona memory.

On AMA-Bench, RSM-full reaches $83\%$ of Full-Context quality at $32\%$ of the token cost at a $4$k prompt budget and outperforms the closest compact-memory baselines in the ${\sim}2.6$k--${\sim}5$k regime. Matched multi-seed ablations show that most of this advantage comes from the merge rule ($+5.7$\,pp over Online K-Means and matched-$\tau$ DP-means) and the grouped packer ($+5.02 \pm 1.00$\,pp over flat concatenation). On RealMem, the same compact-memory pattern reproduces under matched evaluation. Throughout the paper, we treat the merge rule and grouped packer as the main explanation for AMA-Bench, and the $v_1$ retrieval rule only as benchmark-conditional supporting evidence: on BGE-normalized AMA-Bench the retrieval contrast is null once clustering and packing are fixed (Table~\ref{tab:clustering_retrieval_2x2}, A$-$D cell, $p{=}0.40$), whereas on LoCoMo-Plus grouped Cognitive it improves results on $2/4$ relation families.

\paragraph{Related work.}
Recent agent-memory systems attack long-horizon behavior through compression, selective context construction, or structured memory management~\citep{wu2025human2ai,du2026memory,jia2025longterm,hu2024hiagent,zhou2025mem1,xu2025amem}. Related compact-memory systems such as ACON, Latent Context Compilation, Active Context Compression, and Memori focus mainly on \emph{what} to compress or summarize~\citep{kang2025acon,li2026latentctx,verma2026acc,borro2026memori}. Our question is narrower: under a tight prompt budget, which design choices at the \emph{vector-index-and-assembly} layer actually determine answer quality? This makes our closest neighbors online vector-index methods such as threshold-based clustering, Online K-Means, Streaming PCA, and FIFO prototype memories~\citep{kulis2012dpmeans,fritzke1995gng,oja1982simplified,packer2023memgpt}. RSM is complementary to higher-level prompt-management systems~\citep{izacard2023atlas,zhong2024memorybank,maharana2024locomo}; selective forgetting is outside scope.

\begin{figure}[t]
\centering
\vspace{-2mm}
\includegraphics[height=42mm,keepaspectratio]{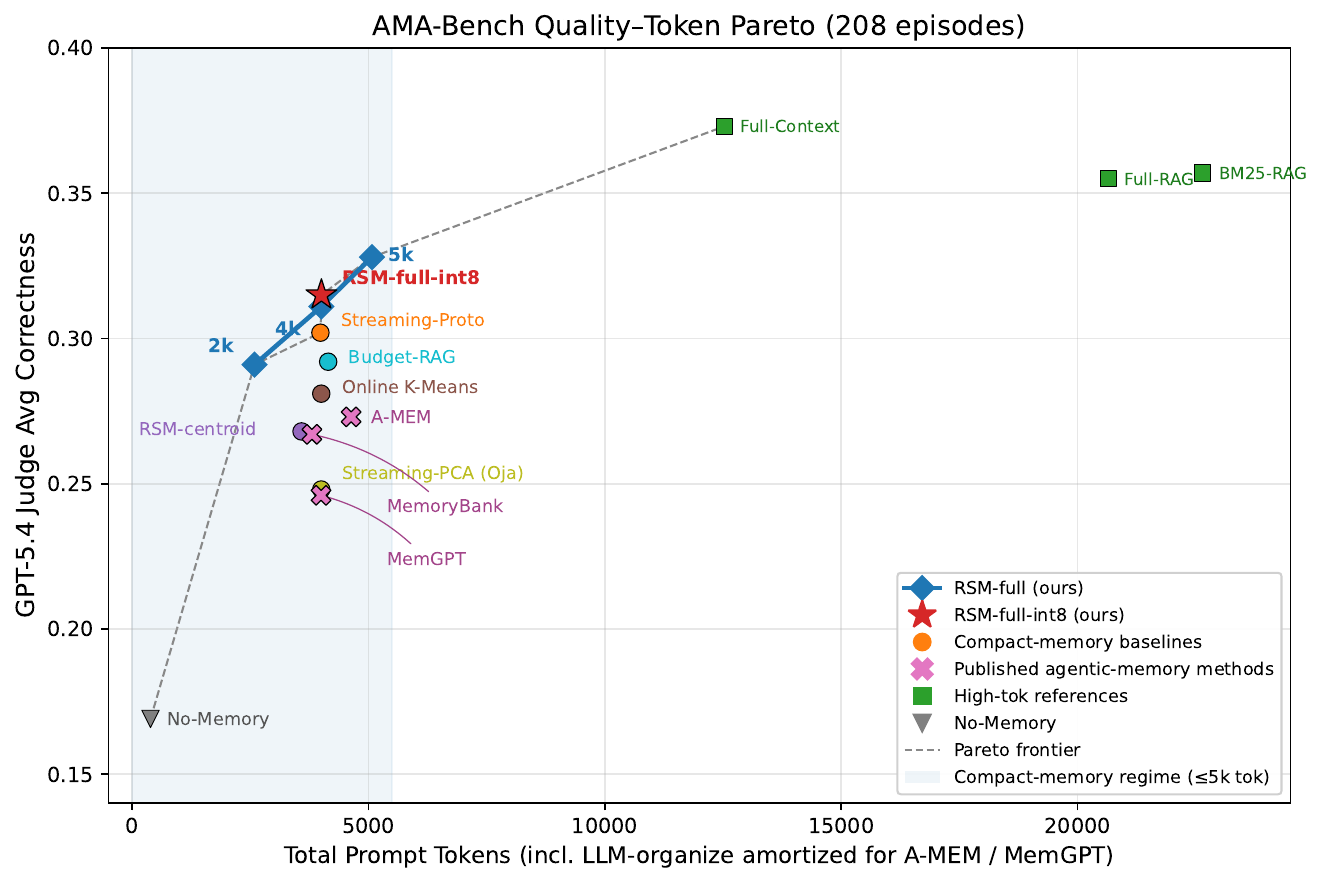}
\vspace{-1mm}
\caption{Quality--token trade-off on AMA-Bench. The shaded region marks the compact-memory regime (${\le}5$k tokens).}
\vspace{-2mm}
\label{fig:ama_pareto}
\end{figure}

\paragraph{Contributions.}
\begin{enumerate}[leftmargin=*,topsep=2pt,itemsep=2pt]
\item \textbf{A strong compact-memory operating point.} On AMA-Bench, RSM-full reaches $83\%$ of Full-Context quality at $32\%$ of the token cost and, under matched four-seed averaging, beats Online K-Means by $+3.5$--$6.0$\,pp across the ${\sim}2.6$k--${\sim}5$k regime (Table~\ref{tab:oja_kmeans_baselines}, Figure~\ref{fig:ama_pareto}).
\item \textbf{A mechanism account for that gain.} On AMA-Bench, matched ablations attribute most of the improvement to the cosine-gated max-member merge rule ($+5.7$\,pp over Online K-Means and matched-$\tau$ DP-means) and the atom-aware grouped packer ($+5.02 \pm 1.00$\,pp over flat concatenation; Tables~\ref{tab:clustering_retrieval_2x2},~\ref{tab:packer_ablation_lean}).
\item \textbf{Replication on RealMem.} The same compact-memory pattern reproduces on an independent long-horizon persona-memory benchmark: RSM-full improves on Budget-RAG, matches BM25-RAG within noise, and outperforms Streaming-Proto and A-MEM under matched evaluation (Table~\ref{tab:realmem}).
\end{enumerate}

We also delimit the scope of these claims. The $v_1$ retrieval rule provides conditional evidence on LoCoMo-Plus grouped Cognitive, but it is null on BGE-normalized AMA-Bench once clustering and packing are fixed; similarly, raw LoCoMo and LongMemEval define boundary regimes in which the compact-memory advantage weakens or disappears.

\section{Method}
\label{sec:method}
This section describes the part of RSM that is needed to explain its quality--token Pareto advantage in the compact-memory regime. The main text therefore focuses on the two components directly supported by the AMA ablations: the cosine-gated write rule that determines how streaming memories are merged, and the atom-aware grouped packer that determines how retrieved content is assembled at prompt time. Higher-rank storage machinery (stored basis $V_m$, coefficient summary $c_m$, Grassmann merge, adaptive rank, int8/PQ quantisation) is retained as a deployment path and deferred to Appendix~\ref{sec:method_storage_extensions}.

\paragraph{Component 1 --- Cosine-gated max-member merge.}
As trajectory chunks arrive one at a time, RSM incrementally routes each chunk $\mathbf{h}$ to an atom (cluster) or allocates a new atom. Let $\mu_m$ denote the running $\ell_2$-normalised mean of atom $m$. The write gate is
\begin{equation}
\label{eq:write_gate}
\begin{aligned}
 m^* &= \arg\max_m \cos(\mathbf{h}, \mu_m),\\[2pt]
 \text{action}(\mathbf{h}) &=
 \begin{cases}
  \text{merge into } m^*, & \text{if } \max\!\Bigl(\cos(\mathbf{h}, \mu_{m^*}),\ \max_i \cos(\mathbf{h}, x_{m^*,i})\Bigr) \ge \tau,\\
  \text{allocate a new atom}, & \text{otherwise.}
 \end{cases}
\end{aligned}
\end{equation}
A single threshold $\tau$ is calibrated per embedding space on ${\le}50$ unlabelled validation examples via $\tau\approx Q_{0.70}$ of pairwise cosines (\S\ref{sec:hyperparams}). The write cost is $\mathcal{O}(Md)$ per chunk ($M$ = live atom count). The key structural choice is the \texttt{max\_mem} check: a new chunk can merge when either the centroid or any existing member exceeds $\tau$. This makes the gate more permissive than pure centroid thresholding and is the axis tested directly against Online K-Means and matched-$\tau$ DP-means in Table~\ref{tab:clustering_retrieval_2x2}.

\paragraph{Component 2 --- Atom-aware grouped context assembly.}
At prompt-construction time, RSM retrieves the top-$K$ atoms, ranks members within each atom, and packs the selected chunks \emph{grouped by atom} (with atom headers and temporal ordering) rather than as a flat rank-ordered list. This grouped packer is the paper's second core component: when clustering and retrieval are held fixed, replacing atom-aware grouping with flat concatenation costs $5.02 \pm 1.00$ percentage points on a matched three-seed AMA ablation (Table~\ref{tab:packer_ablation_lean}).

\paragraph{Retrieval rule and storage variants.}
Each atom $m$ can be scored by
\begin{equation}
 \Score(\mathbf{q}, m) = \lvert \mathbf{v}_{1,m}^\top \mathbf{q}\rvert,
 \label{eq:proj_score}
\end{equation}
where $\mathbf{v}_{1,m}$ is the top singular vector of the atom's member matrix. On $\ell_2$-normalised BGE embeddings, this score is equivalent, for retrieval purposes, to normalised-centroid cosine once the atom-aware packer is held fixed (Table~\ref{tab:clustering_retrieval_2x2}, A$-$D). Higher-rank storage and quantised variants remain useful for cross-geometry deployment (ToolBench, MMLU) and are detailed in Appendix~\ref{sec:method_storage_extensions}.

\paragraph{Complexity.}
Per-query complexity is $\mathcal{O}(Md)$, identical to centroid cosine at one inner product per atom. Storage extensions (int8, PQ, Grassmann merge, adaptive rank) are appendix material because the primary BGE AMA results are explained by clustering and packing rather than by deeper use of the stored basis.

\section{Experimental Setup}

\paragraph{Benchmark roles.} The benchmarks are chosen to test different parts of the paper's central claim. AMA-Bench (208 episodes)~\citep{amabench2025} is the primary compact-memory benchmark and provides the main evidence for the quality--token operating point and the matched ablations on merging and packing. RealMem is the second positive end-to-end benchmark and tests whether the same compact-memory advantage transfers to long-horizon persona memory under matched evaluation. LoCoMo-Plus grouped Cognitive relation-family slices then serve as conditional external evidence for the deployed full pipeline, especially for the benchmark-conditional $v_1$ retrieval result. Raw LoCoMo~\citep{maharana2024locomo} and LongMemEval define boundary regimes in which pure dense retrieval is expected to weaken and the compact-memory advantage may diminish. The MMLU sequential stream serves as an anti-forgetting stress test at matched bytes, and ToolBench (catbal $N{=}2000$, appendix) is a storage-compression reference. Dataset details and operating-point configurations are reported in Appendix~\ref{sec:rsm_config_table}.

\paragraph{Methods.} On AMA-Bench we compare against No-Memory, Full-Context, BM25-RAG, Full-RAG, Budget-RAG, a Streaming-Proto FIFO prototype, MemGPT, MemoryBank, A-MEM, Streaming-PCA (Oja), Online K-Means, and matched-$\tau$ online DP-means. On RealMem we compare against No-Memory, BM25-RAG, Full-RAG, Budget-RAG, Streaming-Proto, RSM-centroid, RSM-full-int8, and the reproduced A-MEM baseline; MemoryBank and MemGPT are omitted from the main table because under our harness both score at or below No-Memory (Appendix~\ref{sec:realmem_repro_excluded}). On LoCoMo-Plus grouped Cognitive we report No-Memory, Full-RAG, Budget-RAG, Streaming-Proto, A-MEM, RSM-cent-retr, and RSM-full.

\paragraph{Hyperparameters and selection protocol.}\label{sec:hyperparams}
All RSM hyperparameters ($\tau$, $k$, \texttt{chunk\_turns}) are selected on held-out validation splits that are strictly disjoint from the reported test sets. The cosine threshold $\tau$ varies with embedding geometry (calibration rule: $\tau \approx Q_{0.70}$ of pairwise cosines from $\leq 50$ unlabelled validation examples): MMLU (all-MiniLM) $\tau{=}0.16$; ToolBench (text-embedding-3-small) $\tau{=}0.65$; AMA-Bench/LoCoMo (BGE-large) $\tau{=}0.85$.

\paragraph{Evaluation protocol.} AMA-Bench uses GPT-4o-mini as the agent (temperature $0$, max-output $256$) and a uniform llmmelon GPT-5.4\footnote{``llmmelon GPT-5.4'' refers to a GPT-5 family model (version 5.4, accessed at temperature $0$ via an aggregator-mediated endpoint) used uniformly for all judge calls in this paper to ensure within-paper consistency. Judge-comparability against the AMA-Bench shipped GPT-4o judge, plus split-half and rerun-stability diagnostics, are reported in Appendix~\ref{sec:judge_comparison}.} rejudge over all 2,496 QAs per method. RealMem uses GPT-4o-mini as the agent and llmmelon GPT-5.4 with the official single-hop judge template, pooled over three stream-permutation seeds. LoCoMo-Plus grouped Cognitive uses GPT-4o-mini as the agent and the official LLM-as-judge protocol with llmmelon GPT-5.4, scored correct $=1$, partial $=0.5$, wrong $=0$. Raw LoCoMo and LongMemEval use reference token-F1. ToolBench uses GPT-4o as both agent and judge, and MMLU / LLaMA hidden-state experiments use dataset-native Recall@1. Judge-stability diagnostics are reported in Appendix~\ref{sec:judge_comparison}.

\section{Main Results}

\paragraph{Unified comparison protocol.}
The main results proceed from AMA-Bench, to RealMem, to conditional external evidence and boundary conditions. All within-table deltas are computed inside a single prompt pipeline on paired QA items, so absolute levels should be compared only within each table. AMA-Bench uses the production v20 pipeline with \texttt{chunk\_turns}=5, \texttt{retrieve\_k}=6, and $\tau{=}0.85$. RealMem uses the same BGE stack with \texttt{chunk\_mode=session}, \texttt{retrieve\_k}=5, and $\tau{=}0.85$ at a $4$k-token budget. LoCoMo-Plus grouped Cognitive uses the same BGE stack with benchmark-fixed settings, without test-set tuning.

\subsection{AMA-Bench: compact-memory operating point}
\label{sec:e2e_agent}
AMA-Bench is the paper's primary benchmark because it most directly tests the compact-memory quality--token frontier. In the ${\sim}2.6$k--${\sim}5$k regime, where practical systems must trade answer quality against prompt budget rather than optimize raw quality at any cost, RSM-full occupies a favorable Pareto operating point.

\paragraph{Main AMA result (Table~\ref{tab:ama_bench_v5}).}
At the matched ${\sim}4$k token budget, RSM-full ($0.311$) outperforms Streaming-Proto ($0.302$) by $+0.9$\,pp and Budget-RAG ($0.292$) by $+1.9$\,pp, delivering $\mathbf{83\%}$ of Full-Context quality at $\mathbf{32\%}$ of the token cost. Higher-token references (Full-Context, Full-RAG, BM25-RAG) remain stronger in absolute quality, but they define the upper-quality region outside the compact-memory regime rather than the frontier at matched budgets.

Paired bootstrap ($10{,}000$ resamples, per-QA, $n{=}2{,}496$) gives Streaming-Proto: $\Delta{=}+1.72$\,pp, $95\%$ CI $[-0.24, +3.69]$, $p{=}0.091$ \emph{NS} (single-seed); Budget-RAG: $\Delta{=}+2.76$\,pp, $95\%$ CI $[+0.76, +4.69]$, $p{=}0.004$. Under $4$-seed averaging on the SP side (Table~\ref{tab:oja_kmeans_baselines}), the RSM-full vs.\ SP gap reaches $+2.68$\,pp ($p{=}0.002$). When published agentic-memory methods are run at the same ${\sim}4$k-token budget under the authors' original algorithms, MemGPT~\citep{packer2023memgpt}, MemoryBank~\citep{zhong2024memorybank}, and A-MEM~\citep{xu2025amem} reach $0.246$, $0.267$, and $0.273\,{\pm}\,0.003$, respectively, all at least $3.8$\,pp below RSM-full-int8 ($0.315$).

For fairness, all rows in this table are evaluated inside the same prompt pipeline, and all are single deterministic runs except A-MEM, which is reported as $\mu{\pm}\sigma$ over three stream-permutation seeds because note generation introduces server-side nondeterminism even at temperature $0$. A-MEM, MemoryBank, and MemGPT also use the same snippet-expansion mechanism as RSM-full, keeping per-query prompt usage within $\pm 5$\% of RSM-full's $4{,}001$ tokens. The RSM-centroid row should be read as a joint retrieval-and-packing control; the isolated retrieval contrast with the atom-aware packer held fixed appears in the A--D comparison of Table~\ref{tab:clustering_retrieval_2x2} and is null ($p{=}.40$). Table~\ref{tab:ama_bench_v5} gives the full per-metric breakdown, while Tables~\ref{tab:oja_kmeans_baselines} and~\ref{tab:packer_ablation_lean} report the symmetric multi-seed analyses for RSM-full.

Within Table~\ref{tab:ama_bench_v5}, Streaming-Proto and Budget-RAG are reported at the matched ${\sim}4$k token budget (calibrated bt$=$4880 / bt$=$6240; raw data \texttt{data/token\_matched/}). The token column reports prompt tokens plus amortized LLM-organization cost, so the Avg/kTok column uses the same total-cost denominator throughout. Under this accounting, RSM-full remains ahead of every compact-memory baseline at the matched ${\sim}4$k budget (Streaming-Proto $0.076$, Budget-RAG $0.070$).

For the reproduced 2025 agentic-memory baselines, we treat A-MEM as the closest harness-compatible comparator in AMA-Bench and RealMem. We keep MemGPT and MemoryBank in Table~\ref{tab:ama_bench_v5} only as same-pipeline reproduced references, not as method-level upper bounds: the same default-hyperparameter reproduction issues discussed for RealMem in Appendix~\ref{sec:realmem_repro_excluded} plausibly affect AMA-Bench too, though less severely because AMA-Bench episodes are shorter.

\begin{table}[t]
\centering
\small
\caption{AMA-Bench end-to-end comparison. Methods above the midline require more than $12$k tokens.}
\label{tab:ama_bench_v5}
\begin{threeparttable}
\setlength{\tabcolsep}{3pt}
\resizebox{\linewidth}{!}{%
\begin{tabular}{lccccccc}
\toprule
Method & Avg (per-ep) & Recall & CausInf & StaUpd & StaAbs & TotalTok$^{\heartsuit}$ & Avg/kTok \\
\midrule
Full-Context & \textbf{0.373} & \textbf{0.430} & 0.369 & \textbf{0.380} & \textbf{0.249} & 12{,}519 & $0.030$ \\
BM25-RAG & 0.357 & 0.405 & 0.366 & 0.363 & 0.234 & 22{,}648 & $0.016$ \\
Full-RAG & 0.355 & 0.415 & \textbf{0.373} & 0.340 & 0.234 & 20{,}642 & $0.017$ \\
\midrule
RSM-full-int8 (ours, $k{=}8$)$^{\ast}$ & \textbf{0.315} & \textbf{0.373} & 0.337 & 0.295 & \textbf{0.196} & 4{,}003 & $\mathbf{0.079}$ \\
\textbf{RSM-full (ours, $k{=}8$)} & 0.311 & 0.361 & 0.342 & \textbf{0.308} & 0.169 & \textbf{4{,}001} & $\mathbf{0.078}$ \\
Streaming-Proto & 0.302 & 0.337 & \textbf{0.347} & 0.280 & 0.203 & 3{,}982 & $0.076$ \\
Budget-RAG & 0.292 & 0.335 & 0.326 & 0.269 & 0.193 & 4{,}146 & $0.070$ \\
Online K-Means ($K_{\max}{=}16$)$^{\S}$ & 0.281 & --- & --- & --- & --- & 4{,}000 & $0.070$ \\
RSM-centroid (ours, centroid control) & 0.268 & 0.302 & 0.305 & 0.250 & 0.172 & 3{,}583 & $0.075$ \\
Streaming-PCA (Oja, $k{=}4$)$^{\S}$ & 0.248 & --- & --- & --- & --- & 4{,}000 & $0.062$ \\
A-MEM$^{\ddagger}$ & $0.273\,{\pm}\,0.003$ & 0.319 & 0.299 & 0.246 & 0.182 & $4{,}218{+}414$ & $0.059$ \\
MemoryBank$^{\ddagger}$ & 0.267 & 0.315 & 0.299 & 0.250 & 0.150 & 3{,}802 & $0.070$ \\
MemGPT$^{\ddagger}$ & 0.246 & 0.274 & 0.285 & 0.232 & 0.152 & $3{,}803{+}193$ & $0.062$ \\
No-Memory & 0.169 & 0.168 & 0.242 & 0.127 & 0.130 & 388 & $0.436$ \\
\bottomrule
\end{tabular}%
}
\begin{tablenotes}[flushleft]
\footnotesize
\item[$^{\ast}$] Int8 vs float32 gap is within noise.
\item[$^{\S}$] Oja/K-Means report the lean-pipeline 4-seed mean.
\item[$^{\heartsuit}$] TotalTok = prompt tokens + amortized LLM-organization cost.
\item[$^{\ddagger}$] Reproduced with the authors' original algorithms inside the same prompt pipeline; see the surrounding text and Appendix~\ref{sec:realmem_repro_excluded} for reproduction limitations.
\end{tablenotes}
\end{threeparttable}
\end{table}

\paragraph{Multi-budget comparison (Table~\ref{tab:oja_kmeans_baselines}).}
Under the lean pipeline with four stream-permutation seeds, RSM-full outperforms Online K-Means at all three tested budgets ($+3.7$/$+3.5$/$+6.0$\,pp at ${\sim}2.6$k/${\sim}4$k/${\sim}5$k, all $p{<}.001$) and Oja by $+4.6$--$9.5$\,pp. The advantage is therefore consistent across the target compact-memory budget range, not confined to a single operating point.

\begin{table}[t]
\centering
\footnotesize
\caption{AMA-Bench multi-seed comparison against the closest vector-index baselines.}
\label{tab:oja_kmeans_baselines}
\begin{minipage}{\linewidth}
\begin{threeparttable}
\setlength{\tabcolsep}{4pt}
\renewcommand{\arraystretch}{1.12}
\begin{tabular*}{\linewidth}{@{\extracolsep{\fill}}lccccc@{}}
\toprule
& \multicolumn{2}{c}{\makecell[c]{Multi-seed mean\\(per-QA)}} & \multicolumn{3}{c}{\makecell[c]{Mean gain vs. RSM-full ref.\\(percentage points)}} \\
\cmidrule(lr){2-3}\cmidrule(lr){4-6}
\makecell[l]{Method} & \makecell[c]{bt=\\2048/4096/8192} & Seeds & $\sim$2.6k & $\sim$4k & $\sim$5k \\
\midrule
\textbf{RSM-full} (ref.) & $.279/.316/.360$ & 4/4/4 & ref & ref & ref \\
\textbf{Oja} ($k{=}4$) & $.232/.248/.265$ & 4/4/3 & $\mathbf{+4.64}$ & $\mathbf{+6.78}$ & $\mathbf{+9.46}$ \\
\textbf{Online K-Means} & $.242/.281/.299$ & 4/4/3 & $\mathbf{+3.65}$ & $\mathbf{+3.50}$ & $\mathbf{+6.04}$ \\
\bottomrule
\end{tabular*}
\begin{tablenotes}[flushleft]
\footnotesize
\item Both baselines use flat assemblers; the assembler confound is controlled in Table~\ref{tab:kmeans_atom_aware}.
\item Entries in the right block report the mean quality gap relative to the RSM-full reference, in percentage points.
\end{tablenotes}
\end{threeparttable}
\end{minipage}
\end{table}

\subsection{Why it works: causal decomposition on AMA-Bench}
\label{sec:component_ablation}

\paragraph{$2{\times}3$ factorial decomposition (Table~\ref{tab:clustering_retrieval_2x2}).}
With the atom-aware packer held fixed, the main result is that the clustering gain comes from RSM's max-member merge rule rather than from generic growing-$K$ behavior or from the retrieval rule. In the six-cell design \{RSM, K-Means, DP-means\} $\times$ \{$|\mathbf{v}_1^\top\mathbf{q}|$, centroid cosine\}, RSM improves over K-Means by $+5.7$\,pp (A$-$B) and over matched-$\tau$ DP-means by $+5.5$\,pp (A$-$E), both $p{<}.0001$. DP-means is indistinguishable from K-Means ($p{=}0.38$), while the retrieval-rule contrast on RSM clusters is null (A$-$D, $p{=}0.40$).

This is a matched clustering ablation: all six cells use the same atom-aware packer, the same per-episode stream permutation, and the same three-seed pooled evaluation ($n{=}7{,}488$). DP-means uses the $\ell_2$ threshold equivalent to the cosine gate $\tau{=}0.85$, namely $\lambda{=}\sqrt{2-2\tau}{=}\sqrt{0.3}{\approx}0.5477$, with $K_{\max}{=}16$ matched to Online K-Means. Significance is computed with paired bootstrap over $10{,}000$ resamples.

\begin{table}[t]
\centering\small
\caption{$2{\times}3$ decomposition of RSM, Online K-Means, and matched-$\tau$ DP-means on BGE-normalized AMA-Bench.}
\label{tab:clustering_retrieval_2x2}
\resizebox{\linewidth}{!}{%
\begin{tabular}{lccc}
\toprule
 & $\lvert\mathbf{v}_{1,m}^\top\mathbf{q}\rvert$ retrieval & Centroid cosine retrieval & Retrieval contrast \\
\midrule
\textbf{RSM (cosine gate, no $K$ cap)} & A: $\mathbf{0.3142}$ & D: $0.3109$ & A$-$D $=+0.33$ (NS, $p{=}.40$) \\
\textbf{Online K-Means ($K_{\max}{=}16$)} & B: $0.2568$ & C: $0.2652$ & B$-$C $=-0.84^{*}$ ($p{=}.015$) \\
\textbf{DP-means ($\lambda{=}\sqrt{0.3}{\approx}0.5477$, $K_{\max}{=}16$)} & E: $0.2597$ & F: $0.2628$ & E$-$F $=-0.30$ (NS, $p{=}.38$) \\
\midrule
\multicolumn{4}{l}{\textbf{Clustering contrasts at $v_1$ retrieval endpoint}:} \\
\quad RSM vs K-Means & \multicolumn{2}{l}{A$-$B $=\mathbf{+5.74^{***}}$\quad CI $[+4.65, +6.84]$, $p{<}.0001$} & \\
\quad \textbf{RSM vs DP-means} & \multicolumn{2}{l}{\textbf{A$-$E $=\mathbf{+5.45^{***}}$\quad CI $[+4.37, +6.53]$, $p{<}.0001$}} & \\
\quad DP-means vs K-Means & \multicolumn{2}{l}{E$-$B $=+0.29$\quad CI $[-0.35, +0.92]$, $p{=}.38$ (NS)} & \\
\multicolumn{4}{l}{\textbf{Clustering contrasts at centroid retrieval endpoint}:} \\
\quad RSM vs K-Means & \multicolumn{2}{l}{D$-$C $=\mathbf{+4.57^{***}}$\quad CI $[+3.55, +5.62]$, $p{<}.0001$} & \\
\quad \textbf{RSM vs DP-means} & \multicolumn{2}{l}{\textbf{D$-$F $=\mathbf{+4.81^{***}}$\quad CI $[+3.79, +5.84]$, $p{<}.0001$}} & \\
\quad DP-means vs K-Means & \multicolumn{2}{l}{F$-$C $=-0.25$\quad CI $[-0.87, +0.40]$, $p{=}.43$ (NS)} & \\
\bottomrule
\multicolumn{4}{l}{\footnotesize Combined A$-$C: $+4.90$\,pp ($p{<}.0001$). Interaction: $+1.17$\,pp ($p{=}.026$).}\\
\end{tabular}
}%
\end{table}

\paragraph{Atom-aware packer ablation (Table~\ref{tab:packer_ablation_lean}).}\label{sec:packer_ablation_lean}
With clustering and retrieval held fixed, the main result is that the atom-aware grouped packer adds $+5.02 \pm 1.00$\,pp over flat concatenation across three seeds. This shows that prompt assembly is not a cosmetic implementation detail but a substantive part of the compact-memory Pareto gain.

This is a matched packer ablation on the full 208-episode AMA test set: Table~\ref{tab:packer_ablation_lean} uses the same RSM clusters, the same $|\mathbf{v}_1^\top q|$ retrieval rule, and the same bt${=}4096$ setup as cell A of Table~\ref{tab:clustering_retrieval_2x2}; only the packer changes.

\begin{table}[ht]
\centering\small
\caption{Three-seed atom-aware packer ablation on the lean AMA-Bench pipeline.}
\label{tab:packer_ablation_lean}
\resizebox{\linewidth}{!}{%
\begin{tabular}{lcccc}
\toprule
Seed & Atom-aware ($k{=}8$) & Flat ($k{=}8$) & $\Delta$ (pp) & Note \\
\midrule
seed 43 & $0.3141$ ($784/2496$) & $0.2600$ ($649/2496$) & $\mathbf{+5.41}$ & Single-seed bootstrap CI $[+3.53,+7.29]$, $p{<}.0001$ \\
seed 44 & $0.3189$ ($796/2496$) & $0.2612$ ($652/2496$) & $\mathbf{+5.77}$ & Matched-protocol replication \\
seed 45 & $0.3097$ ($773/2496$) & $0.2708$ ($676/2496$) & $\mathbf{+3.89}$ & Matched-protocol replication \\
\midrule
\textbf{3-seed mean} & $0.3142$ & $0.2640$ & $\mathbf{+5.02 \pm 1.00}$ & $t$-CI $[+2.54, +7.50]$; $p{<}.0001$ per seed \\
\bottomrule
\multicolumn{5}{l}{\footnotesize Atom-aware numbers are the 3-seed RSM-full results reported in Table~\ref{tab:oja_kmeans_baselines}.}\\
\end{tabular}
}%
\end{table}

\paragraph{Supporting bridge: K-Means with the same packer (Table~\ref{tab:kmeans_atom_aware}).}
As a bridge check, giving K-Means the same atom-aware packer still leaves RSM-full ahead by $+4.90$\,pp ($p{<}.0001$). This is consistent with the cleaner clustering-only contrast in Table~\ref{tab:clustering_retrieval_2x2} and shows that the advantage is not eliminated by equalizing prompt assembly.

\begin{table}[t]
\centering
\small
\caption{K-Means + atom-aware packer ablation on AMA-Bench.}
\label{tab:kmeans_atom_aware}
\setlength{\tabcolsep}{4pt}
\renewcommand{\arraystretch}{1.12}
\resizebox{\linewidth}{!}{%
\begin{tabular}{@{}lccccc@{}}
\toprule
\makecell[l]{Method} & Seed 43 & Seed 44 & Seed 45 & Pool. & $p$ \\
\midrule
\makecell[l]{\textbf{RSM-full}\\($v_1$)} & $0.3141$ & $0.3189$ & $0.3097$ & $\mathbf{0.3142}$ & --- \\
\makecell[l]{K-Means\\(centroid)} & $0.2700$ & $0.2580$ & $0.2676$ & $0.2652$ & --- \\
\midrule
$\Delta$ (RSM $-$ KM, pp) & $+4.40$ & $+6.08$ & $+4.21$ & $\mathbf{+4.90}$ & $\mathbf{<.0001}$ \\
$95\%$ CI & $[+2.56, +6.29]$ & $[+4.17, +7.93]$ & $[+2.32, +6.09]$ & $[+3.79, +5.98]$ & \\
\midrule
\multicolumn{6}{@{}l}{\textit{Domain breakdown (pooled; RSM $-$ K-Means)}} \\
OPENWORLD-QA ($n{=}1080$) & & & & $\mathbf{+10.27}$ & $\mathbf{<.0001}$ \\
WEB ($n{=}1116$) & & & & $\mathbf{+8.51}$ & $\mathbf{<.0001}$ \\
TEXT2SQL ($n{=}1836$) & & & & $\mathbf{+8.32}$ & $\mathbf{<.0001}$ \\
SOFTWARE ($n{=}1296$) & & & & $\mathbf{+4.18}$ & $\mathbf{.002}$ \\
EMBODIED-AI ($n{=}1080$) & & & & $-2.68$ & $.042$ ($\alpha$-low) \\
Game ($n{=}1080$) & & & & $-1.58$ & $.259$ NS \\
\bottomrule
\end{tabular}
}%
\vspace{1mm}
{\footnotesize Three stream-permutation seeds $\{43,44,45\}$, pooled $n{=}7{,}488$ paired QA. Both arms use the same atom-aware packer at a 4096-token budget. The observed $+4.90$\,pp is therefore a joint clustering-and-retrieval effect. The clustering-only contrast with retrieval held fixed is A$-$B in Table~\ref{tab:clustering_retrieval_2x2}.}
\end{table}

\paragraph{Mechanism check: compression behavior at longer horizon.}\label{par:mechanism_long_horizon}
We treat the following as a mechanism check only --- not as a long-horizon quality claim --- because at $N{=}32$k AMA-Long Track-A accuracy sits at the No-Memory ceiling for all methods, so any quality $\Delta$ at this horizon speaks to bank-size compression rather than to end-to-end answer quality (Appendix~\ref{sec:horizon_scaling}). Within that scope, on AMA-Long Track-A super-trajectories (AMA-Bench episodes stitched at $N{=}16$k and $N{=}32$k turns, $75$ episodes per $N$), removing the max-member fallback (\textbf{RSM-cent-merge}; a write-side ablation distinct from the retrieval-side ``RSM-centroid'' of Tables~\ref{tab:ama_bench_v5} and~\ref{tab:realmem}) inflates atom count by $+8.84\%$ / $+10.56\%$ (paired $t{>}21$, $p{<}10^{-32}$ at each $N$) at null paired quality cost (pooled $\Delta{=}{+}0.22$\,pp, $p{=}.70$ NS). The merge rule's compression behavior therefore extends beyond AMA-Bench's per-episode horizon, even though the long-horizon quality envelope is unchanged.

\subsection{A second positive compact-memory benchmark: RealMem}
\label{sec:realmem}
RealMem is a long-horizon persona-memory benchmark where the agent answers first-person questions about a character after streaming $135$--$276$ conversational sessions per persona ($\sim$50k--100k tokens of dialogue). We evaluate on \textbf{all $10$ released personas} (\texttt{adeleke\_okonjo}, \texttt{ethan\_hunt}, \texttt{kenta\_tanaka}, \texttt{kim\_ji\_young}, \texttt{liam\_o\_connor}, \texttt{lin\_wanyu}, \texttt{oliver\_smith}, \texttt{pak\_budi}, \texttt{sarah\_miller}, \texttt{sophie\_dubois}) with $114$--$168$ paper-released QA pairs per persona ($1{,}415$ unique QA total), at three stream-permutation seeds $\{43,44,45\}$ matching the AMA protocol (pooled $n{=}4{,}243$). RSM-full uses the same production settings as LoCoMo-Plus: $\tau{=}0.85$, $k{=}8$, \texttt{chunk\_mode=session}, \texttt{retrieve\_k=5}, budget $4$k tokens, agent GPT-4o-mini, judge llmmelon GPT-5.4 with the official single-hop template.

\paragraph{Main result (Table~\ref{tab:realmem}).}
RealMem is the paper's second positive end-to-end result alongside AMA-Bench. It reproduces the same compact-memory pattern on an independent benchmark. RSM-full is the top-scoring compact-memory method and \emph{strictly Pareto-dominates} Budget-RAG (higher accuracy, $+0.69$\,pp, with fewer prompt tokens) while matching BM25-RAG's accuracy at slightly fewer tokens (NS). It also significantly outperforms Streaming-Proto ($+2.97$\,pp, $p{<}.001$) and RSM-centroid ($+2.08$\,pp, $p{<}.001$). \textbf{A-MEM is the closest reproduced 2025 agentic-memory comparator at $+1.65$\,pp ($p{<}.001$).} The reproduced MemoryBank and MemGPT runs fall at or below the No-Memory floor under our harness, suggesting reproduction limitations on RealMem rather than a meaningful method-level comparison; we are investigating and have therefore omitted those rows from Table~\ref{tab:realmem}, with details in Appendix~\ref{sec:realmem_repro_excluded}. Cross-seed std is $\le 0.013$ for every reported RSM-full and baseline row, consistent with the order-invariance pattern in Table~\ref{tab:order_invariance}.

\begin{table}[t]
\centering\small
\caption{RealMem results on the full 10-persona set. $\Delta$ denotes RSM-full minus baseline.}
\label{tab:realmem}
\resizebox{\linewidth}{!}{%
\begin{tabular}{lcccc}
\toprule
Method & 3-seed mean Acc & TotalTok$^{\heartsuit}$ & $\Delta$ vs RSM-full (pp) & $p$-value \\
\midrule
\textbf{RSM-full} ($k{=}8$, ours) & $\mathbf{0.4684}$ & $3{,}705$ & --- & --- \\
BM25-RAG & $0.4657$ & $3{,}773$ & $+0.27$ & $0.47$ NS \\
Budget-RAG & $0.4615$ & $3{,}737$ & $\mathbf{+0.69^{\ast\ast}}$ & $\mathbf{0.006}$ \\
Full-RAG & $0.4569$ & $3{,}513$ & $\mathbf{+1.15^{\ast\ast}}$ & $\mathbf{0.002}$ \\
\textbf{A-MEM}$^{\dagger}$ & $0.4519$ & $3{,}753{+}316$ & $\mathbf{+1.65^{\ast\ast\ast}}$ & $\mathbf{<.001}$ \\
RSM-full-int8 & $0.4506$ & $3{,}513$ & $\mathbf{+1.78^{\ast\ast\ast}}$ & $\mathbf{<.001}$ \\
RSM-centroid ($k{=}1$) & $0.4477$ & $3{,}515$ & $\mathbf{+2.08^{\ast\ast\ast}}$ & $\mathbf{<.001}$ \\
Streaming-Proto & $0.4388$ & $3{,}512$ & $\mathbf{+2.97^{\ast\ast\ast}}$ & $\mathbf{<.001}$ \\
No-Memory & $0.2317$ & $0$ & $+23.67^{\ast\ast\ast}$ & $<.001$ \\
\bottomrule
\multicolumn{5}{l}{\footnotesize $^{\heartsuit}$TotalTok includes amortized LLM organization cost. $^{\dagger}$ marks reproduced baseline.}\\
\end{tabular}
}%
\end{table}

\paragraph{Interpretation.}
RealMem differs from AMA-Bench in task structure and lexical content, but the qualitative conclusion is the same: compact-memory performance is strongest when clustering and grouped assembly preserve chunk-level detail under a tight token budget. The effect size is smaller than the AMA $+4.90$\,pp RSM-vs-K-Means gap, plausibly because RealMem has weaker atom structure and stronger lexical cues, but the direction is unchanged. We therefore treat RealMem as the paper's second positive end-to-end benchmark and as external validation of the compact-memory Pareto claim.

\subsection{Conditional external evidence and boundary conditions}
LoCoMo-Plus grouped Cognitive provides conditional external evidence for the deployed full pipeline, while raw LoCoMo and LongMemEval define the boundary regimes where the clustered-memory advantage weakens.

\paragraph{LoCoMo-Plus grouped Cognitive: conditional external evidence (Table~\ref{tab:locomoplus_grouped}).}
Across the four relation families, RSM-full with $v_1$ retrieval is significantly ahead of the $v_1$-free RSM-centroid baseline on \texttt{causal} and \texttt{goal}, while \texttt{state} and \texttt{value} families are null. LoCoMo-Plus therefore provides conditional external evidence that the retrieval-rule contrast can activate on a grouped multi-cue benchmark, on 2 of 4 relation families. A matched-qid flat-vs-grouped control in Appendix Table~\ref{tab:locomoplus_flat_vs_grouped} is consistent with that interpretation.

\begin{table}[t]
\centering
\small
\caption{LoCoMo-Plus grouped Cognitive across all four relation families.}
\label{tab:locomoplus_grouped}
\begin{minipage}{\linewidth}
\begin{threeparttable}
\setlength{\tabcolsep}{4pt}
\renewcommand{\arraystretch}{1.12}
\resizebox{\linewidth}{!}{%
\begin{tabular}{@{}lcccc@{}}
\toprule
\makecell[l]{Method / slice} & \makecell[c]{Causal\\($n{=}101$)} & \makecell[c]{State\\($n{=}100$)} & \makecell[c]{Goal\\($n{=}100$, new)} & \makecell[c]{Value\\($n{=}100$, new)} \\
\midrule
\textbf{RSM-full} ($v_1$, $k{=}8$) & $\mathbf{0.396}$ / $3446$ & $0.240$ / $3913$ & $\mathbf{0.210}$ / $398$ & $0.310$ / $4042$ \\
\midrule
RSM-cent-retr ($v_1$-free) & $0.327$ / $2319^{*}$ & $0.250$ / $3128$ & $0.190$ / $314$ & $0.370$ / $3894^{\dagger}$ \\
Full-RAG & $0.347$ / $2319$ & $0.180$ / $3128^{*}$ & $0.180$ / $314$ & $0.370$ / $3894^{\dagger}$ \\
Budget-RAG & $0.327$ / $3536^{*}$ & $0.250$ / $3701$ & $0.150$ / $2765^{\dagger}$ & $0.310$ / $3894$ \\
Streaming-Proto & $0.327$ / $2319^{*}$ & $0.260$ / $3128$ & $0.150$ / $314^{\dagger}$ & $0.350$ / $3894^{\dagger}$ \\
\textcolor{gray!70}{A-MEM$^{\ddagger}$} & \textcolor{gray!70}{$0.149$ / $3783$} & \textcolor{gray!70}{$0.110$ / $3879$} & \textcolor{gray!70}{$0.050$ / $3435$} & \textcolor{gray!70}{$0.270$ / $3144$} \\
No-Memory & $0.218$ / $0$ & $0.140$ / $0$ & $0.120$ / $0$ & $0.170$ / $0$ \\
\bottomrule
\end{tabular}
}%
\begin{tablenotes}[flushleft]
\footnotesize
\item Entries report quality (avg) / mean prompt tokens.
\item Superscripts denote paired McNemar significance vs.~RSM-full (ties dropped): $^{*}p{<}.05$, $^{**}p{<}.01$, $^{***}p{<}.001$, $^{\dagger}$ NS.
\item Exact $p$-values by relation family are reported in Appendix Table~\ref{tab:locomoplus_flat_vs_grouped}.
\item $^{\dagger}$ Budget-RAG on \texttt{state} is the only non-significant exception; $^{\ddagger}$ A-MEM is excluded because of configuration mismatch.
\item RSM-cent-retr shares RSM-full's clusters and atom-aware packer but swaps $|v_1^\top q|$ for centroid-cosine retrieval (retrieval-axis ablation).
\end{tablenotes}
\end{threeparttable}
\end{minipage}
\end{table}

\paragraph{Boundary regimes.}
On raw LoCoMo and LongMemEval, pure-dense methods underperform BM25/Full-RAG and RSM-full collapses toward RSM-centroid, consistent with the low-$\alpha$ operating envelope. Positive end-to-end claims are therefore scoped to AMA-Bench, RealMem, and the grouped LoCoMo-Plus slices, while raw LoCoMo and LongMemEval serve as boundary conditions.

\section{Discussion}
\label{sec:discussion}
The evidence supports a simple hierarchy of claims. AMA-Bench establishes the main compact-memory operating-point result, RealMem reproduces the same pattern on an independent benchmark, and the matched AMA ablations explain the gain mainly through the max-member merge rule and the atom-aware grouped packer. The $v_1$ retrieval rule remains conditional supporting evidence only: it is null on BGE-normalized AMA-Bench once clustering and packing are fixed, but activates on $2/4$ LoCoMo-Plus grouped relation families.

Practically, RSM-full is most relevant when answer quality must be maintained under roughly 2--5k prompt tokens and the stream contains enough repeated latent structure to form multi-member atoms. When that structure is weak, the method converges toward centroid-equivalent behavior and lexical or hybrid retrieval can become preferable; Table~\ref{tab:atom_diagnostics} should therefore be read only as a descriptive cross-benchmark diagnostic. ToolBench and MMLU suggest that the same code path remains useful under matched-byte storage constraints and anti-forgetting stress tests, but higher-token methods remain stronger above $12$k tokens.

\section{Limitations}
\label{sec:limitations}
Positive end-to-end claims are anchored by AMA-Bench and RealMem under a single BGE embedding stack; LoCoMo-Plus grouped Cognitive provides only conditional supporting evidence. End-to-end experiments use one embedder and one primary judge stack, so cross-embedder replication remains future work. RSM also does not support selective forgetting, and keyword-heavy settings can still favor lexical or hybrid retrieval. Finally, the long-horizon mechanism-extension ablation (§\ref{par:mechanism_long_horizon}, Appendix~\ref{sec:horizon_scaling}) is run on AMA-Long Track-A super-trajectories stitched from AMA-Bench episodes, so it isolates bank-size compression rather than an independent long-horizon quality claim.

\section*{Societal Impact}
RSM reduces per-agent memory footprint but still requires safeguards against cross-session leakage, demographic-bucket clustering, and judge-amplified errors.

\section{Conclusion}
RSM-full defines a strong quality--token Pareto point in the compact-memory regime. Across AMA-Bench and RealMem, the main empirical pattern is consistent: under tight prompt budgets, compact-memory performance depends mainly on how streaming memories are merged and how retrieved content is assembled. On AMA-Bench, most of the gain comes from two simple design choices---the cosine-gated max-member merge rule and the atom-aware grouped packer---and on RealMem the same advantage reproduces under matched evaluation. RSM-full is therefore most useful when answer quality must be preserved under roughly $2$k--$5$k prompt tokens; outside that regime, lexical retrieval or higher-token systems can remain preferable. Future work includes selective forgetting, cross-embedder replication, and broader multi-seed validation of appendix-level checks.

\bibliographystyle{plain}
\bibliography{references}

@inproceedings{lewis2020rag,
  title={Retrieval-Augmented Generation for Knowledge-Intensive NLP Tasks},
  author={Lewis, Patrick and Perez, Ethan and Piktus, Aleksandra and Petroni, Fabio and Karpukhin, Vladimir and Goyal, Naman and K{"u}ttler, Heinrich and Lewis, Mike and Yih, Wen-tau and Rockt{"a}schel, Tim and Riedel, Sebastian and Kiela, Douwe},
  booktitle={Advances in Neural Information Processing Systems},
  year={2020}
}

@inproceedings{yao2023react,
  title={ReAct: Synergizing Reasoning and Acting in Language Models},
  author={Yao, Shunyu and Zhao, Jeffrey and Yu, Dian and Du, Nan and Shafran, Izhak and Narasimhan, Karthik and Cao, Yuan},
  booktitle={International Conference on Learning Representations},
  year={2023}
}

@article{gao2023retrieval,
  title={Retrieval-Augmented Generation for Large Language Models: A Survey},
  author={Gao, Yunfan and Xiong, Yue and Gao, Xinyu and Jia, Kang and Pan, Jinliu and Bi, Yuntian and Dai, Yansong and Sun, Jiawei and Wang, Haofen and others},
  journal={arXiv preprint arXiv:2312.10997},
  year={2023}
}

@article{maharana2024locomo,
  title={Evaluating Very Long-Term Conversational Memory of {LLM} Agents},
  author={Maharana, Adyasha and Lee, Dong-Ho and Tulyakov, Sergey and Bansal, Mohit and Barbieri, Francesco and Fang, Yuwei},
  journal={arXiv preprint arXiv:2402.17753},
  year={2024}
}

@inproceedings{izacard2023atlas,
  title={Atlas: Few-shot Learning with Retrieval Augmented Language Models},
  author={Izacard, Gautier and Lewis, Patrick and Lomeli, Maria and Hosseini, Lucas and Petroni, Fabio and Schick, Timo and Dwivedi-Yu, Jane and Joulin, Armand and Riedel, Sebastian and Grave, Edouard},
  booktitle={Journal of Machine Learning Research},
  volume={24},
  pages={1--43},
  year={2023}
}

@inproceedings{zhong2024memorybank,
  title={MemoryBank: Enhancing Large Language Models with Long-Term Memory},
  author={Zhong, Wanjun and Guo, Lianghong and Gao, Qiqi and Ye, He and Wang, Yanlin},
  booktitle={Proceedings of the AAAI Conference on Artificial Intelligence},
  year={2024}
}

@article{packer2023memgpt,
  title={{MemGPT}: Towards {LLM}s as Operating Systems},
  author={Packer, Charles and Fang, Vivian and Patil, Shishir G. and Lin, Kevin and Wooders, Sarah and Gonzalez, Joseph E.},
  journal={arXiv preprint arXiv:2310.08560},
  year={2023}
}

@article{xu2025amem,
  title={{A-MEM}: Agentic Memory for {LLM} Agents},
  author={Xu, Wujiang and Liang, Zujie and Chen, Jintao and Zhu, Tiantian and Gao, Xingjun and Nie, Yuanhao and Tang, Yongfeng},
  journal={arXiv preprint arXiv:2502.12110},
  year={2025}
}

@article{amabench2025,
  title={{AMA-Bench}: Evaluating Long-Horizon Memory for Agentic Applications},
  author={Zhao, Yujie and Yuan, Boqin and Huang, Junbo and Yuan, Haocheng and Yu, Zhongming and Xu, Haozhou and Hu, Lanxiang and Shankarampeta, Abhilash and Huang, Zimeng and Ni, Wentao and Tian, Yuandong and Zhao, Jishen},
  journal={arXiv preprint arXiv:2602.22769},
  year={2026}
}

@article{hu2024hiagent,
  title={{HiAgent}: Hierarchical Working Memory Management for Solving Long-Horizon Agent Tasks with Large Language Model},
  author={Hu, Mengkang and Chen, Tianxing and Chen, Qiguang and Mu, Yao and Shao, Wenqi and Luo, Ping},
  journal={arXiv preprint arXiv:2408.09559},
  year={2024}
}

@article{zhou2025mem1,
  title={{MEM1}: Learning to Synergize Memory and Reasoning for Efficient Long-Horizon Agents},
  author={Zhou, Zijian and Qu, Ao and Wu, Zhaoxuan and Kim, Sunghwan and Prakash, Alok and Rus, Daniela and Zhao, Jinhua and Low, Bryan Kian Hsiang and Liang, Paul Pu},
  journal={arXiv preprint arXiv:2506.15841},
  year={2025}
}

@article{hendrycks2021mmlu,
  title={Measuring Massive Multitask Language Understanding},
  author={Hendrycks, Dan and Burns, Collin and Basart, Steven and Zou, Andy and Mazeika, Mantas and Song, Dawn and Steinhardt, Jacob},
  journal={Proceedings of the International Conference on Learning Representations (ICLR)},
  year={2021}
}

@article{oja1982simplified,
  title={Simplified Neuron Model as a Principal Component Analyzer},
  author={Oja, Erkki},
  journal={Journal of Mathematical Biology},
  volume={15},
  number={3},
  pages={267--273},
  year={1982},
  publisher={Springer}
}

@inproceedings{kulis2012dpmeans,
  title={Revisiting k-means: New Algorithms via Bayesian Nonparametrics},
  author={Kulis, Brian and Jordan, Michael I.},
  booktitle={Proceedings of the 29th International Conference on Machine Learning (ICML)},
  year={2012}
}

@article{fritzke1995gng,
  title={A Growing Neural Gas Network Learns Topologies},
  author={Fritzke, Bernd},
  journal={Advances in Neural Information Processing Systems},
  volume={7},
  year={1995}
}

@article{wu2025human2ai,
  title={From Human Memory to AI Memory: A Survey on Memory Mechanisms in the Era of {LLM}s},
  author={Wu, Yaxiong and Liang, Sheng and Zhang, Chen and Wang, Yichao and Zhang, Yongyue and Guo, Huifeng and Tang, Ruiming and Liu, Yong},
  journal={arXiv preprint arXiv:2504.15965},
  year={2025}
}

@inproceedings{jia2025longterm,
  title={Evaluating the Long-Term Memory of Large Language Models},
  author={Jia, Zixi and Liu, Qinghua and Li, Hexiao and Chen, Yuyan and Liu, Jiqiang},
  booktitle={Findings of the Association for Computational Linguistics: {ACL} 2025},
  pages={19759--19777},
  year={2025}
}

@article{du2026memory,
  title={Memory for Autonomous {LLM} Agents: Mechanisms, Evaluation, and Emerging Frontiers},
  author={Du, Pengfei},
  journal={arXiv preprint arXiv:2603.07670},
  year={2026}
}

@article{kang2025acon,
  title={{ACON}: Optimizing Context Compression for Long-horizon {LLM} Agents},
  author={Kang, Minki and Chen, Wei-Ning and Han, Dongge and Inan, Huseyin A. and Wutschitz, Lukas and Chen, Yanzhi and Sim, Robert and Rajmohan, Saravan},
  journal={arXiv preprint arXiv:2510.00615},
  year={2025}
}

@article{li2026latentctx,
  title={Latent Context Compilation: Distilling Long Context into Compact Portable Memory},
  author={Li, Zeju and Zhou, Y. and Xu, Q.},
  journal={arXiv preprint arXiv:2602.21221},
  year={2026}
}

@article{verma2026acc,
  title={Active Context Compression: Autonomous Memory Management in {LLM} Agents},
  author={Verma, Nikhil},
  journal={arXiv preprint arXiv:2601.07190},
  year={2026}
}

@article{borro2026memori,
  title={Memori: A Persistent Memory Layer for Efficient, Context-Aware {LLM} Agents},
  author={Borro, Luiz C. and Macarini, Luiz A. B. and Tindall, Gordon and Montero, Michael and Struck, Adam B.},
  journal={arXiv preprint arXiv:2603.19935},
  year={2026}
}

\appendix
\section{Supplementary Tables and Controls}

\paragraph{Variant definitions.}
The main paper uses \textbf{RSM-full} as the algorithmic instance, \textbf{RSM-full-int8} as the storage-compressed deployment variant, and \textbf{RSM-centroid} as a centroid-only control. The table below is retained for terminology consistency.

\begin{table}[ht]
\centering
\small
\caption{Atom-membership diagnostics across benchmarks.}
\label{tab:atom_diagnostics}
\setlength{\tabcolsep}{4pt}
\resizebox{\linewidth}{!}{%
\begin{tabular}{lccccll}
\toprule
Benchmark & Atoms & $\alpha$ (Basis\%) & AvgSize & $\Delta$ & Control & Judge \\
\midrule
\multicolumn{7}{l}{\emph{High-$\alpha$ regime ($\alpha{\ge}50\%$): clustered-memory advantage activates}} \\
ToolBench-catbal ($N{=}2{,}000$) & 227 & $56.8\%$ & 8.8 & $+0.25$\,pp & RSM-centroid & GPT-4o \\
AMA-Bench (208\,ep) & 2.3/ep & $62\%^\dagger$ & 10.2 & $+4.3$\,pp & RSM-centroid & GPT-5.4 \\
\midrule
\multicolumn{7}{l}{\emph{Intermediate regime (partially untested)}} \\
ToolBench-hard ($N{=}500$) & 282 & $29.8\%$ & 1.8 & $+1.8$\,pp (int8) & RSM-centroid & GPT-4o \\
\midrule
\multicolumn{7}{l}{\emph{Low-$\alpha$ regime ($\alpha{\le}5\%$): RSM-full $\approx$ centroid, graceful degradation}} \\
LoCoMo (10\,conv) & 38.4 & ${\le}5\%$ & ${\sim}2.6$ & $+0.58$\,pp & RSM-centroid & Token F1 \\
LongMemEval ($N{=}500$) & 47.6 & $1.2\%$ & 1.05 & $-0.3$\,pp & RSM-centroid & Token F1 \\
\midrule
\multicolumn{7}{l}{\emph{High-$\alpha$, different control (supporting evidence)}} \\
MMLU ($N{=}1{,}383$) & 31 & ${\sim}80\%$ & 44.6 & $+1.8$\,pp$^\ddagger$ & Streaming-Proto & Rule-based \\
\bottomrule
\end{tabular}
}%

\vspace{1mm}
{\footnotesize $^\dagger$AMA-Bench reports both atom-level and episode-level proxies because each episode contains only ${\sim}2.3$ atoms on average.
$^\ddagger$MMLU uses a sequential-stream protocol and a different control baseline.}
\end{table}

\begin{table}[ht]\centering\footnotesize
\caption{RSM pipeline variants.}
\label{tab:variant_defs}
\setlength{\tabcolsep}{4pt}
\begin{tabular}{lcccc}
\toprule
Name & Rank ($k$) & Retrieval & Context builder & Storage/atom \\
\midrule
\textbf{RSM-full} & 2--8 & $|\mathbf{v}_1^\top\mathbf{q}|$ & atom-aware & $(k{+}1)d$ floats \\
\textbf{RSM-full-int8} & 2--8 & $|\mathbf{v}_1^\top\mathbf{q}|$ & atom-aware & $(k{+}1)d$ int8 \\
RSM-centroid & 0 & $\cos(\mathbf{q},\mu_m)$ & basic (flat) & $d$ floats \\
\bottomrule
\end{tabular}
\end{table}

\paragraph{Matched-qid flat-vs-grouped control for LoCoMo-Plus.}
This control is useful for interpretation and rebuttal, but it is secondary to the main AMA and grouped-LoCoMo tables, so we report it here for completeness.

\begin{table}[ht]
\centering
\small
\caption{Flat-vs-grouped matched-qid control on LoCoMo-Plus (bal40, $n{=}40$).}
\label{tab:locomoplus_flat_vs_grouped}
\begin{minipage}{\linewidth}
\begin{threeparttable}
\setlength{\tabcolsep}{4pt}
\renewcommand{\arraystretch}{1.12}
\resizebox{\linewidth}{!}{%
\begin{tabular}{@{}lccccc@{}}
\toprule
Protocol & \makecell[c]{RSM-full\\($v_1$, $k{=}8$)} & RSM-cent-retr & Budget-RAG & \makecell[c]{Streaming-\\Proto} & Full-RAG \\
\midrule
flat      & $0.175$ / $3100$ & $0.150$ / $3106$ & $\mathbf{0.225}$ / $3875$ & $0.150$ / $3260$ & $0.100$ / $3189$ \\
grouped   & $\mathbf{0.200}$ / $1914$ & $0.100$ / $2650$ & $\mathbf{0.225}$ / $3852$ & $0.200$ / $2565$ & $0.175$ / $2752$ \\
\midrule
$\Delta$(grouped$-$flat) & $+0.025$ / $-1187$ & $-0.050$ / $-456$ & $+0.000$ / $-23$ & $+0.050$ / $-696$ & $+0.075$ / $-437$ \\
\bottomrule
\end{tabular}
}%
\begin{tablenotes}[flushleft]
\footnotesize
\item Entries report mean quality / mean prompt tokens.
\item Paired comparison (RSM-full vs.~RSM-cent-retr): flat $\Delta{=}{+}2.5$\,pp ($p{=}1.00$); grouped $\Delta{=}{+}10.0$\,pp ($p{=}0.22$).
\item At $n{=}40$, the grouped effect is directional rather than conclusive.
\item Budget-RAG remains at $0.225$ under both protocols.
\end{tablenotes}
\end{threeparttable}
\end{minipage}
\end{table}

\section{Additional Experiments and Diagnostics}
\label{sec:appendix_additional}

\subsection{Method: Storage Extensions (stored basis, coefficient summary, merge update, adaptive rank, quantisation)}
\label{sec:method_storage_extensions}

This subsection collects higher-rank storage machinery that is supplementary to the primary BGE-AMA benchmark. On MMLU sequential, $k{\in}\{2,4,8\}$ are all equivalent in quality (Table~\ref{tab:mmlu_seq}), with RSM-full-int8 at $k{=}2$ as the lowest-byte operating point. The higher-rank code path is retained for cross-geometry continuity and for deployment with streams of unknown dimensionality.

\paragraph{Per-atom stored basis $V_m$ and coefficient summary $c_m$.} Each atom $m$ stores an orthonormal basis $V_m\in\mathbb{R}^{d\times k}$ (top-$k$ right singular vectors of $X_m$) and $c_m = \frac{1}{n_m}\sum_i X_m[i,:] V_m \in\mathbb{R}^k$. Main-paper retrieval Eq.~\ref{eq:proj_score} uses $V_m$'s first column only.

\paragraph{Bounded-buffer exact SVD (the evaluated system).} Each atom maintains up to \texttt{MAX\_MEMBER\_VECS}$=20$ buffered members; exact SVD is recomputed at each merge. Storage accounting distinguishes \emph{serialised} (persistent $V_m + \mu_m$) from \emph{peak resident} (transient workspace): peak resident overhead for AMA-Bench is $\approx 188$\,KB per episode, released after buffer flush. All matched-byte comparisons use serialised storage.

\paragraph{Online Grassmann merge (scalability path).} When $n_m$ exceeds the buffer, basis updates use an approximate weighted Fréchet mean on $\mathcal{G}(d,k)$ via QR + eigendecomposition in $O(dk^2)$: $B = w_i (Q^\top V_i)(Q^\top V_i)^\top + w_n (Q^\top V_n)(Q^\top V_n)^\top$, $V_i' = Q\,\mathrm{eig}_k(B)$ with $Q$ the QR basis of $[\sqrt{w_i}V_i,\sqrt{w_n}V_n]$. The $k$-truncation provides self-correction. Most AMA-Bench atoms accumulate $\le 20$ members, so Grassmann updates are exercised only rarely.

\paragraph{Coefficient update.} $c_i' = (n_i c_i + c_n)/(n_i+1)$; task-utility-weighted variant uses reward weight $w(\rho){=}\exp(\rho/\tau_w)$.

\paragraph{Adaptive rank selection.} For open-domain streams, $k^*(m) = \min\{k: \sum_{i=1}^k \sigma_{m,i}^2 / \sum_j \sigma_{m,j}^2 \ge \eta\}$ with $\eta{=}0.95$, $k_{\min}{=}16$, $k_{\max}{=}128$. Evaluated benchmarks use fixed rank $k{\in}\{1,2,4,8\}$ per operating point for controlled comparability.

\paragraph{Basis quantisation.} int8 per-column min-max gives $4\times$ smaller $V_m$ with $+0.06$\,pp end-to-end quality on AMA-Bench (Table~\ref{tab:judge_comparison}, RSM-full-int8); PQ gives $\sim 8\times$. Per-query complexity remains $\mathcal{O}(Md)$.

\subsection{Algorithm E: Subspace-Projected Per-Chunk Scorer (Negative Result)}
\label{sec:alg_e}

\paragraph{Motivation.} The main paper's retrieval rule $|\mathbf{v}_1^\top\mathbf{q}|$ only uses one direction of the stored $k$-column basis $V_m$. One might reasonably ask whether a denser rule that uses $V_m$ more fully---akin to full subspace projection $\|V_m^\top\mathbf{q}\|_2$ but applied at the chunk level rather than the atom level---could extract empirical advantage from the $k{-}1$ otherwise-unused columns, and thereby empirically justify higher-rank storage under BGE-normalised embeddings.

\paragraph{Algorithm.} Let $V_m\in\mathbb{R}^{d\times k}$ be atom $m$'s stored orthonormal basis (columns are the top-$k$ mean-centred right singular vectors of the member matrix). Let $\mathbf{x}_i$ denote the $i$-th member embedding and $\mathbf{q}$ the query. Algorithm~E keeps the exact RSM write gate, the exact $|\mathbf{v}_{1,m}^\top\mathbf{q}|$ atom-level filter, and the exact atom-aware packer (\S\ref{sec:method}). The only change is the \emph{per-member scoring} function used inside the retrieve/pack step: ambient cosine $\mathbf{x}_i^\top\mathbf{q}$ is replaced by the subspace-projected inner product
\begin{equation}
 \text{score}_{\text{E}}(\mathbf{x}_i,\mathbf{q}\mid m) = \mathbf{x}_i^\top V_m V_m^\top \mathbf{q}
 = \sum_{j=1}^{k} (\mathbf{v}_{j,m}^\top \mathbf{x}_i)(\mathbf{v}_{j,m}^\top \mathbf{q}).
 \label{eq:alg_e_app}
\end{equation}
Every summand $j{\ge}2$ contributes only when the query and the candidate chunk are both aligned with a non-principal singular direction of the atom, so this score is the minimal change that could in principle let $k{>}1$ basis columns pay off in scoring. RSM-full (rank-8 storage, $k{=}8$, ambient cosine) and Algorithm~E (rank-8 storage, $k{=}8$, subspace-projected) are tested under the same pipeline; the only difference is the per-chunk scoring formula.

\paragraph{Protocol.} Full 208-episode AMA-Bench ($n{=}2{,}496$ paired QA). Agent gpt-4o-mini; judge gpt-5.4. Chunking, budget, and packer match Table~\ref{tab:kmeans_atom_aware}'s lean pipeline. This is a \emph{single-seed} run, not pooled with the three-seed $\{43,44,45\}$ cell A; absolute baseline differs by ${\sim}0.4$\,pp (within the stream-permutation envelope), but paired $\Delta$'s are within this run. Two paired conditions: RSM-full (ambient cosine) and Algorithm~E (subspace-projected, $k{=}8$). Paired bootstrap ($10{,}000$ resamples).

\paragraph{Result.} Table~\ref{tab:alg_e_result} reports the full $n{=}2{,}496$ paired comparison. The result is negative: subspace-projected per-chunk scoring does not outperform the standard RSM scorer, reinforcing the paper's main reading that the causal gains on BGE come from clustering and packing rather than from deeper use of the stored basis during scoring.
\begin{table}[ht]
\centering\small
\caption{Algorithm~E vs. RSM-full on the full 208-episode AMA-Bench.}
\label{tab:alg_e_result}
\begin{threeparttable}
\resizebox{\linewidth}{!}{%
\begin{tabular}{lcccc}
\toprule
Method & Mean score (correct/$n$) & $\Delta$ vs RSM-full & $95\%$ CI & $p$ \\
\midrule
\textbf{RSM-full} ($k{=}8$, ambient cosine, ref) & $0.3185$ ($795/2{,}496$) & (ref) & --- & --- \\
Algorithm~E ($k{=}8$, subspace-projected) & $0.3109$ ($776/2{,}496$) & $-0.76$ & $[-2.12, +0.60]$ & $0.27$ \\
\bottomrule
\end{tabular}
}%
\begin{tablenotes}[flushleft]
\footnotesize
\item Per-domain breakdown (Algorithm~E $-$ RSM-full): WEB $+1.01$\,pp ($p{=}.56$, high-$\alpha$), OPENWORLD\_QA $+0.79$ ($p{=}.74$), Game $-0.91$, SOFTWARE $-1.26$, EMBODIED\_AI $-1.48$, and TEXT2SQL $-1.89$ ($p{=}.15$). No domain reaches $p{<}.05$.
\item Raw data: \texttt{data/rsm\_multidir\_v2\_full208.json}. Bootstrap script: \texttt{scripts/experiment/multidir\_v2\_paired\_bootstrap.py}.
\end{tablenotes}
\end{threeparttable}
\end{table}

\paragraph{Interpretation.} On BGE-normalised embeddings, the $|\mathbf{v}_1^\top\mathbf{q}|$ atom filter already selects chunks within a tight within-cluster subspace, so ambient cosine approximately equals the projected score; the extra $j{\ge}2$ terms act as re-weighting rather than added signal. The small negative $\Delta$ ($-0.76$\,pp, CI $[-2.12, +0.60]$) is under single-seed power and does not distinguish ``small signal loss'' from ``null fluctuation''. The reading is that RSM's structural novelty on BGE lives in clustering + packing, not in deeper scoring-level use of the stored basis.

\subsection{RSM Operating Points per Section}
\label{sec:rsm_config_table}

\begin{table}[ht]
\centering
\small
\caption{RSM operating points by experimental section.}
\label{tab:rsm_config}
\begin{threeparttable}
\resizebox{\linewidth}{!}{%
\begin{tabular}{p{3.2cm}p{2.8cm}p{2.2cm}p{2.4cm}p{3.0cm}}
\toprule
Section & Stored object & Retrieval rule & Write gate & Charged storage \\
\midrule
\S\ref{sec:real_baselines} (LLaMA h.s.) & Full subspace $V_m$ $+$ centroid ($k{\in}\{2,4\}$) & Proj.\ $|\mathbf{v}_1^\top q|$ (Eq.~\ref{eq:proj_score}) & Cosine $\tau{=}0.65$ & $(dk{+}d)\!\times\!4$\,B (f32); int8 optional \\
Appendix~\ref{sec:latency} (Latency) & Full subspace $V_m$ (col-major) & Proj.\ precomputed $V_m^\top q$ & N/A (isolated profiling) & $dk\!\times\!4$\,B \\
\S\ref{sec:mmlu_domain} (MMLU seq.) & Full subspace $V_m$ $+$ centroid ($k{=}2$) & Proj.\ $|\mathbf{v}_1^\top q|$ (Eq.~\ref{eq:proj_score}) & Cosine $\tau{=}0.16$ & $(2d{+}d)\!\times\!4$\,B (f32); int8 optional \\
\S\ref{sec:e2e_agent} (AMA main table) & Full subspace $V_m$ $+$ centroid ($k{=}8$) & Proj.\ $|\mathbf{v}_1^\top q|$ & Cosine $\tau{=}0.85$ & $(dk{+}d)\!\times\!4$\,B (f32); int8 optional \\
\S\ref{sec:e2e_agent} (LoCoMo main table) & Full subspace $V_m$ $+$ centroid ($k{=}8$) & Proj.\ $|\mathbf{v}_1^\top q|$ (pure dense) & Cosine $\tau{=}0.85$ & $(dk{+}d)\!\times\!4$\,B (f32); int8 optional \\
\bottomrule
\end{tabular}
}%
\begin{tablenotes}[flushleft]
\footnotesize
\item All rows instantiate the same RSM framework. We use the cosine gate as the deployed default because a residual-energy gate ($g{=}1.2$) matches AMA-Bench quality at matched pipeline but over-splits atoms on high-dimensional orthogonal embeddings (Table~\ref{tab:judge_comparison}).
\item Theory/noise sections and the sequential MMLU stress test use the full-subspace instance; centroid rows are deployment-restricted controls. Full-subspace int8 preserves near-centroid byte cost while retaining projection retrieval. $^\flat$\,Setting~C is a synthetic storage-scaling demonstration rather than a headline quality benchmark.
\end{tablenotes}
\end{threeparttable}
\end{table}

\subsection{Retrieval Latency}\label{sec:latency}

Table~\ref{tab:latency} reports median query latency (p50) for RSM-full versus Full-RAG cosine search at varying memory sizes $M$ (number of stored atoms / vectors), profiled on a single CPU (no GPU). \textbf{RSM-full-Opt} uses a precomputed subspace basis ($V_m$ stored column-major); \textbf{RSM-full-Naive} recomputes the projection matrix $P_{V_m} = V_m V_m^\top$ per query. The main experiments use the top-1 score $|\mathbf{v}_{1,m}^\top\mathbf{q}|$, which has the same $\mathcal{O}(Md)$ scaling as Full-RAG cosine. To provide a conservative upper bound, the table instead profiles the more expensive full-basis score $\|V_m^\top\mathbf{q}\|_2$. Even under this upper bound, RSM-full-Opt is 10--20$\times$ faster than the naive implementation and adds only 0.15--0.37\,ms over Full-RAG at $M{\le}50$, far below typical LLM inference latency.

\begin{table}[t]
\centering
\small
\caption{Median query latency (ms) as a function of memory size $M$.}
\label{tab:latency}
\begin{threeparttable}
\resizebox{\linewidth}{!}{%
\begin{tabular}{rcccc}
\toprule
$M$ & Full-RAG (ms) & RSM-full-Naive (ms) & RSM-full-Opt (ms) & Speedup (Naive/Opt) \\
\midrule
 10 & 0.030 & 3.390 & 0.176 & $19{\times}$ \\
 20 & 0.033 & 1.213 & 0.340 & $ 4{\times}$ \\
 50 & 0.085 & 3.354 & 0.367 & $ 9{\times}$ \\
100 & 0.060 & 10.997 & 9.889 & $ 1.1{\times}$ \\
200 & 4.481 & 14.775 & 18.699 & --- \\
\bottomrule
\end{tabular}
}%
\begin{tablenotes}[flushleft]
\footnotesize
\item At $M{=}100$, cache effects reduce RSM-full-Naive overhead; at $M{=}200$, cache and memory-bandwidth effects dominate.
\end{tablenotes}
\end{threeparttable}
\end{table}

At $M{\le}50$ (high-overlap workloads where RSM produces 10--50 atoms), the full-basis variant adds only $0.15$--$0.37$\,ms per query---small compared to typical LLM inference latency ($\ge100$\,ms per forward pass). \textbf{The top-1 score $|\mathbf{v}_{1,m}^\top\mathbf{q}|$ used in all main experiments requires only a single inner product per atom ($\mathcal{O}(d)$), matching Full-RAG cosine search in both theoretical and practical latency.} Table~\ref{tab:latency} profiles the full-basis $\|V_m^\top\mathbf{q}\|_2$ variant ($\mathcal{O}(Mkd)$) as an upper bound for potential future multi-direction scoring variants; at $M{\ge}200$ the full-basis variant exceeds Full-RAG latency and is not recommended for that regime, but the deployed top-1 scorer remains $\mathcal{O}(Md)$ throughout. RSM-full-Naive is $10$--$20{\times}$ slower than RSM-full-Opt because it rebuilds $P_{V_m}$ ($\mathcal{O}(kd^2)$) on every query; precomputing basis columns is the critical optimization.

\section{Notation Reference}
\label{sec:notation}
\begin{table}[ht]
\centering
\small
\caption{Key notation used throughout the paper.}
\label{tab:notation}
\resizebox{\linewidth}{!}{%
\begin{tabular}{ll}
\toprule
Symbol & Description \\
\midrule
$H \in \mathbb{R}^{n \times d}$ & Trajectory matrix ($n$ time steps, hidden dimension $d$) \\
$V_m \in \mathbb{R}^{d \times k}$ & Subspace basis matrix of atom $m$ (orthonormal columns) \\
$P_{V_m} = V_m V_m^\top$ & Orthogonal projection onto {\rm span}$(V_m)$ \\
$c_m \in \mathbb{R}^{k}$ & Coefficient summary vector of atom $m$ \\
$e_m = V_m c_m$ & Reconstruction vector (legacy evaluation proxy only) \\
$\mathcal{G}(k, d)$ & Grassmann manifold of $k$-planes in $\mathbb{R}^d$ \\
$\tau$ & Merge threshold (cosine similarity gate for new trajectory) \\
$\tau_G$ & Grassmann gate threshold for subspace fusion \\
$\eta$ & Spectral energy threshold for adaptive rank selection ($\eta{=}0.95$) \\
$k^*$ & Adaptively selected subspace rank from singular-value energy spectrum \\
$\delta(m;h)$ & Projection residual: $\|h\|_2^2 - \|P_{V_m}h\|_2^2$ \\
$\alpha$ & Importance weight in weighted merge update \\
$M$ & Current number of atoms in the memory bank \\
$N$ & Total number of trajectories processed (sequence length) \\
$C$ & Number of task categories in evaluation \\
$\Delta$ & Minimum inter-centroid $\ell_2$ separation \\
$\sigma$ & Per-trajectory isotropic noise level (synthetic experiments) \\
$\TopOneProj$ & Top-1 retrieval quality: subspace projection score $|\mathbf{v}_{1,m^*}^\top\mathbf{q}|$ \\
$\TopOneCos$ & Top-1 retrieval quality: cosine-on-reconstruction (deprecated baseline) \\
$\mathrm{QAME}$ & Quality-Adjusted Memory Efficiency: Compression Ratio $\times$ Top1Proj \\
$\mathrm{Recall}@k$ & Fraction of queries whose true atom appears in the top-$k$ results \\
\bottomrule
\end{tabular}
}%
\end{table}

\section{AMA-Bench: Official GPT-4o Judge vs.\ GPT-5.4 Re-Judge}
\label{sec:judge_comparison}

For cross-paper comparability, Table~\ref{tab:judge_comparison} reports the AMA-Bench compact-memory ranking under both the benchmark's shipped GPT-4o binary judge and our uniform GPT-5.4 re-judge session, \textbf{across all three tested budgets}. The ranking among compact-memory methods is \textbf{identical under both judges at all $3$ budget levels}: RSM-int8 $>$ RSM-full $>$ baselines at ${\le}4$k; RSM-full $>$ RSM-int8 $>$ baselines at ${\sim}5$k. Full-Context's absolute score differs across judges, reflecting known judge-model sensitivity, but the \emph{within-compact-memory} ordering---the basis of our quality ranking---is stable across judges and budgets.

\paragraph{GPT-5.4 re-judge split-half reliability and rerun stability.}
Because our headline numbers come from an author-run GPT-5.4 re-judge session rather than the shipped GPT-4o judge, we verify its internal reliability with two diagnostics at the bt${=}$4k operating point.
(a)~\textbf{Split-half ranking stability.} Splitting the 208-episode set into odd and even episodes (104/104) under the single GPT-5.4 session, RSM-full beats Streaming-Proto by $+2.64$\,pp on the odd half and $+3.69$\,pp on the even half, and beats Budget-RAG by $+4.97$\,pp and $+6.81$\,pp respectively. The ranking RSM-full~$>$~SP~$>$~BR is preserved on every half and the margin is of the same order of magnitude as the full-set margin, so the headline ordering is not an artifact of one difficult sub-sample.
(b)~\textbf{Per-episode rerun correlation.} Across two independent GPT-5.4 re-runs of RSM-full (same agent outputs, independent judge sessions), the per-episode score vectors correlate at Pearson $r{=}0.876$ (208 episodes), with run-level means $0.318$ and $0.320$---consistent with the $3{\times}$ rerun std of $0.003$ reported in the main text. Under the paper's primary judge, per-episode scores are therefore stable to within roughly $\pm0.04$ at the episode level and roughly $\pm0.003$ at the full-test aggregate.

\begin{table}[ht]
\centering\footnotesize
\caption{AMA-Bench compact-memory ranking under the official GPT-4o judge and a GPT-5.4 re-judge.}
\label{tab:judge_comparison}
\setlength{\tabcolsep}{4pt}
\begin{tabular}{l cc cc cc}
\toprule
& \multicolumn{2}{c}{bt=2k} & \multicolumn{2}{c}{bt=4k} & \multicolumn{2}{c}{bt=8k} \\
\cmidrule(lr){2-3}\cmidrule(lr){4-5}\cmidrule(lr){6-7}
Method & 4o & 5.4 & 4o & 5.4 & 4o & 5.4 \\
\midrule
RSM-full-int8 & $\mathbf{.250}$ & $\mathbf{.291}$ & $\mathbf{.342}$ & $\mathbf{.315}$ & $.276$ & $.322$ \\
RSM-full & $.241$ & $.288$ & $.331$ & $.311$ & $\mathbf{.285}$ & $\mathbf{.323}$ \\
\midrule
Streaming-Proto & $.099$ & $.125$ & $.229$ & $.287$ & $.144$ & $.174$ \\
RSM-centroid & $.125$ & $.157$ & $.227$ & $.268$ & $.161$ & $.187$ \\
Budget-RAG & $.084$ & $.111$ & $.207$ & $.259$ & $.133$ & $.168$ \\
\bottomrule
\end{tabular}
\end{table}

\section{Reproducibility}
\label{sec:reproducibility}
All experiments are reproducible from code and result files at the project repository.
Key files include:
\begin{itemize}[leftmargin=1.2em,itemsep=0.2em,topsep=0.3em]
\item \path{ama_bench_full_all_baselines_20260320.json}, \path{ama_bench_fullsub_rsmfulv20_all208_20260325p.json}, \path{data/token_matched/rejudge_gpt54_rsm_full_bt4096_rep2.json}, \path{data/token_matched/rejudge_gpt54_rsm_full_int8_bt4096_rep2.json}, \path{data/token_matched/sp_bt4880.json}, and \path{data/token_matched/br_bt6240.json} (AMA-Bench all-baselines, matched-token baselines, and RSM-full main-table results for Table~\ref{tab:ama_bench_v5}).
\item \path{data/rsm_multi_seed/rsm_full_bt{2048,4096,8192}_seed{42-45}_*.json} and \path{data/rejudged_exp1234/exp34_{oja,online_kmeans}_bt{2048,4096,8192}_seed{42-45}.json} (multi-seed AMA budget sweep for Table~\ref{tab:oja_kmeans_baselines}). \emph{Disclosure}: the bt${=}8192$ sub-cell for Oja and Online K-Means is a 3-seed sub-cell (seed${\in}\{42,43,44\}$); the seed${=}45$ run at bt${=}8192$ was not completed in the release artifact, so the 4-seed mean for that single budget collapses to a 3-seed mean for those two baselines. All other budget$\times$method cells are full 4-seed.
\item \path{kmeans_atom_aware_full208.json} and \path{kmeans_full208_paired_bootstrap.json} (Table~\ref{tab:kmeans_atom_aware}).
\item \path{locomo_public_baselines_legacy_a050_all10_20260327.json} and \path{locomo_public_fullsub_legacy_k8_a050_all10_20260327.json} (raw LoCoMo boundary test).
\item \path{parallel_benchmark_20260414/locomoplus_cognitive_{causal,state,goal,value}_full_grouped_allbaselines_judged_llmmelon_gpt54.json} (LoCoMo-Plus grouped Cognitive 4 families, Table~\ref{tab:locomoplus_grouped}).
\item \path{parallel_benchmark_20260414/locomoplus_cognitive_bal40_{flat,grouped}_allbaselines_judged_llmmelon_gpt54.json} (flat vs grouped matched-qid robustness control, Table~\ref{tab:locomoplus_flat_vs_grouped}).
\item \path{data/ama_noise_n30_sigma08.json} (end-to-end noise robustness probe; $n{=}30$ AMA-Bench episodes $\times$ $3$ methods $\times$ $4$ $\sigma$ levels $\times$ $2$ seeds $=8{,}640$ judged QAs; Table~\ref{tab:noise_robustness}, Figure~\ref{fig:noise_robustness_retention}).
\item \path{data/ama_adversarial_maxcos_strict_n30_ct5.json} (query-aware max-cosine adversarial injection probe, strict top-$K$; $n{=}30$ AMA-Bench episodes $\times$ $3$ methods $\times$ $4$ $K$ levels $\times$ $2$ seeds $=8{,}640$ judged QAs; Table~\ref{tab:adversarial_robustness}, Figure~\ref{fig:adversarial_retention}).
\item \path{llm_eval_v10_catbal_n2000.json} (ToolBench supporting reference).
\item \path{mmlu_seq_fullsub_best_tau016_k2.json} and \path{mmlu_seq_fullsub_sweep.json} (MMLU sequential supporting reference).
\end{itemize}
Additional archived experiments are documented in \path{experiments_archive.tex} alongside their result files.
All experiments use random seed 42 unless noted.


\section{Supporting Cross-Geometry References (moved from main body)}

\subsection*{§A.1 Real LLaMA Hidden States (matched-byte storage compression)}
\label{sec:real_baselines}

We evaluate RSM-full on \emph{real} $d{=}4096$ LLaMA-3.1-8B hidden states from ToolBench ($N{=}2000$, 20 API categories, 80/20 split, 5 seeds). Metric: \textbf{Category Recall@1}. Fair comparison: per-seed byte budget from natural RSM-full bank converted to the largest float32 vector budget. RSM rows use $|\mathbf{v}_1^\top\mathbf{q}|$ retrieval with $\tau{=}0.65$.

\begin{table}[ht]\centering\small
\caption{ToolBench real hidden-state comparison ($N{=}2000$, $d{=}4096$; mean over 5 seeds). $^\ddagger$Offline.}
\label{tab:real_baselines}
\resizebox{\linewidth}{!}{%
\begin{tabular}{lcccccc}
\toprule
Setting & RSM-full ($|\mathbf{v}_1^\top q|$) & Budget-RAG & Stream-Proto & KMeans$^\ddagger$ & Storage & Compression \\
\midrule
Full-RAG (ref.) & $1.000$ & --- & --- & --- & 25{,}600\,KB & $1\times$ \\
\midrule
k\!=\!4, float32 & $\mathbf{0.999}$ & $0.996$ & $0.999$ & $1.000$ & 2{,}624\,KB & $9.8\times$ \\
k\!=\!4, int8 & $\mathbf{0.999}$ & $0.948$ & $0.996$ & $1.000$ & 1{,}064\,KB & $24.1\times$ \\
k\!=\!2, float32 & $\mathbf{0.999}$ & $0.982$ & $0.999$ & $1.000$ & 1{,}616\,KB & $15.8\times$ \\
k\!=\!2, int8 & $\mathbf{0.999}$ & $0.890$ & $0.974$ & $0.996$ & \textbf{812\,KB} & $\mathbf{31.5\times}$ \\
\bottomrule
\end{tabular}
}%
\end{table}

At this operating point, the $k{=}2$ int8 row is the only clearly non-saturated contrast. RSM-full retains $99.9\%$ Recall@1 at $812$\,KB ($31.5\times$ compression vs Full-RAG's $25.6$\,MB), beating same-byte Budget-RAG by $+11.0$\,pp and Streaming-Proto by $+2.6$\,pp. Offline RAG-KMeans is effectively tied with RSM-full at matched bytes (1.000 vs 0.999; $\Delta{-}0.1$\,pp is within sampling noise). The claim in this section is therefore confined to the online streaming-clustered comparison set (Budget-RAG and Streaming-Proto), not to offline batch-clustering methods.

\begin{table}[ht]\centering\small
\caption{Multi-budget comparison on ToolBench LLaMA-3.1-8B hidden states.}
\label{tab:toolbench_sweep}
\resizebox{\linewidth}{!}{%
\begin{tabular}{rcccc}
\toprule
Budget/Compression & \textbf{RSM-centroid} & Budget-RAG & Stream-Proto & KMeans$^\ddagger$ \\
\midrule
$72.2$ ($22.2\times$) & $\mathbf{1.000}$ & $0.970$ & $0.995$ & $0.998$ \\
$54.8$ ($29.2\times$) & $\mathbf{1.000}$ & $0.902$ & $0.998$ & $0.998$ \\
$34.0$ ($47.1\times$) & $\mathbf{0.998}$ & $0.816$ & $0.946$ & $0.996$ \\
$27.4$ ($58.4\times$) & $\mathbf{0.949}$ & $0.759$ & $0.881$ & $0.993$ \\
$24.0$ ($66.7\times$) & $\mathbf{0.938}$ & $0.706$ & $0.845$ & $0.994$ \\
\bottomrule
\end{tabular}
}%
\end{table}

\subsection*{§A.2 MMLU Academic QA (sequential anti-forgetting stress test)}
\label{sec:mmlu_domain}

MMLU~\citep{hendrycks2021mmlu} ($N{=}1{,}383$, 10 STEM/social-science disciplines), sentence-transformers \path{all-MiniLM-L6-v2} ($d{=}384$), global mean-centering + $\ell_2$-normalise. Sequential subject-order stream; vector baselines charged the same bytes as the row-matching RSM-$k$ natural count. Across the rank sweep, RSM-full at $k\in\{2,4,8\}$ reaches the same $0.824$ Recall@1 with full 10/10 subject coverage, while RSM-full-int8 at $k{=}2$ is the lowest-byte operating point at $88.5$\,KB. The zero standard deviation reflects the deterministic RSM write gate under fixed subject ordering; seed variation comes only from baseline KMeans initialization or reservoir sampling.

\begin{table}[ht]\centering\small
\caption{Subject Recall@1 under a sequential topic stream on MMLU. Mean$\pm$std is over 5 seeds for RSM-full at $k{=}2$. $^\ddagger$Offline.}
\label{tab:mmlu_seq}
\resizebox{\linewidth}{!}{%
\begin{tabular}{lccc}
\toprule
Method & Storage & Recall@1 ($\uparrow$) & Coverage ($\uparrow$) \\
\midrule
Full-RAG & 1657.5\,KB & $0.849$ & 10/10 \\
\midrule
Budget-RAG ($k{=}2$ bytes) & 174.0\,KB & $0.273$ & $1/10$ \\
Streaming-Proto ($k{=}2$ bytes) & 174.0\,KB & $0.806$ & 10/10 \\
RAG-KMeans$^\ddagger$ ($k{=}2$ bytes) & 174.0\,KB & $0.831$ & 10/10 \\
\midrule
\textbf{RSM-full} ($\tau{=}0.16$, $k{=}2$) & 174.0\,KB & $\mathbf{0.824 \pm 0.000}$ (5 seeds) & \textbf{10/10} \\
RSM-full ($k{=}4$) & 277.5\,KB & $0.824$ & 10/10 \\
RSM-full ($k{=}8$) & 475.5\,KB & $0.824$ & 10/10 \\
\textbf{RSM-full-int8} ($k{=}2$) & \textbf{88.5\,KB} & $\mathbf{0.824}$ & \textbf{10/10} \\
\bottomrule
\end{tabular}
}%
\end{table}

RSM-full at $k{\in}\{2,4,8\}$ all reach $82.4\%$ Recall@1 with 10/10 subject coverage; increasing $k$ above $2$ adds storage without improving quality. RSM-full at $k{=}2$ is $+1.8$\,pp over Streaming-Proto and $+55.1$\,pp over Budget-RAG (Budget-RAG degenerates to 1/10 coverage). RSM-full-int8 at $k{=}2$ preserves the same quality at $88.5$\,KB (lowest byte point). Offline RAG-KMeans ($0.831$) slightly exceeds RSM-full ($0.824$) by $0.7$\,pp at matched bytes. The claim here is therefore specific to the online/streaming comparison ($+1.8$\,pp vs Streaming-Proto), not to offline batch-clustering methods.

\subsection*{§A.3 Component ablation + scoring variants (summary)}

Main-paper \S\ref{sec:component_ablation} gives the summary numbers. Block 1 (ablation pipeline, $n{=}2496$): $-$SVD merge NS ($-0.76$\,pp, $p{=}.16$); $-$proj$\to$full-basis $-1.80$\,pp ($p{=}.010$); $-$proj$\to\cos(q,V_mc_m)$ $-1.92$\,pp ($p{=}.006$). Block 2 (v20, 164-ep): $-$atom-aware$\to$flat concat $-6.60$\,pp ($p{<}.001$), sign-and-magnitude-consistent with the 3-seed $+5.02\pm1.00$\,pp packer ablation. Extended retrieval scoring variants on 20 AMA episodes (directional sweep, $|t|{<}2$): $|\mathbf{v}_1^\top\mathbf{q}|$ is the recommended default; hybrid and adaptive-$k$ are NS; full-basis $-2.1$\,pp.

\subsection*{§A.4 Additional diagnostics: rank-$4$ null, atom diagnostics, $v_1$ centred vs non-centred (moved from main body)}

\paragraph{Rank-$4$ RSM-full single-seed null check.}
\label{sec:k4_rank_null}
We run RSM-full at $k{=}4$ on seed 43 full 208 episodes at bt${=}4096$, paired against the existing $k{=}8$ seed-43 result. \emph{Result}: $\Delta(k{=}4{-}k{=}8) = -0.64$\,pp, $p{=}.34$ NS; the two rank points are statistically indistinguishable at single-seed power on BGE AMA. MMLU matched-byte sweep (Table~\ref{tab:mmlu_seq}) extends the rank-invariance to $k{\in}\{2,4,8\}$ at $0.824$ Recall@1.

\paragraph{Tier-0 atom diagnostics.}
\label{sec:atom_diagnostics_tier0}
On full 208-ep AMA-Bench (3 seeds, pooled $\sim3268$ atoms): within-cluster cos-to-centroid mean $0.9796$, std $0.0058$ (clusters span $\sim 11.6^\circ$ angular spread on unit sphere); multi-member-atom ratio $0.4215$; $\langle \mathbf{v}_1^{\text{centred}}, \hat\mu\rangle$ mean $0.0484$ (mean-centred $v_1$ is essentially orthogonal to centroid); $\langle \mathbf{v}_1^{\text{non-centred}}, \hat\mu\rangle$ mean $1.0000$ (non-mean-centred $v_1 \equiv \hat\mu$ on tight clusters). Raw: \path{data/atom_diagnostics_tier0.json}.

\paragraph{$v_1^{\text{centred}}$ vs $v_1^{\text{non-centred}}$ vs centroid retrieval.}
\label{sec:v1_centred_vs_noncentred}
Seed 43 full 208, $n{=}2{,}496$. Accuracy: $v_1^{\text{centred}}$ $0.3053$; $v_1^{\text{non-centred}}$ $0.3117$ ($\Delta{+}0.65$ NS $p{=}.27$); $\cos(q,\hat\mu)$ $0.3101$ ($\Delta{+}0.47$ NS $p{=}.45$). All three retrieval scores statistically indistinguishable on BGE-tight clusters. Extending to 3-seed AMA $n{=}7{,}488$ and RealMem 10-persona 3-seed $n{=}4{,}243$, centred and non-centred SVD are empirically indistinguishable ($\Delta{=}-0.36$\,pp NS on AMA; $\Delta{=}-0.24$\,pp NS on RealMem, paired McNemar $p{=}.11$). We therefore retain the classical mean-centred SVD formulation throughout the paper.

\subsection*{§A.5 End-to-end noise robustness: RSM-full, Streaming-Proto, Budget-RAG under chunk-side Gaussian noise}
\label{sec:noise_robustness}

\paragraph{Protocol.} Domain-balanced AMA-Bench subset ($n{=}50$ episodes). BGE chunk embeddings perturbed by i.i.d.\ Gaussian $\mathcal{N}(0,\sigma)$ \emph{before} clustering/indexing and re-normalised; queries clean. $\sigma \in \{0.0, 1.0, 2.0, 3.0\}$ (an extended stress-test range relative to the original $\sigma{\le}0.8$ setting), 2 noise seeds, 3 methods (RSM-full $k{=}8$ $\tau{=}0.85$; Streaming-Proto FIFO $K{=}16$; Budget-RAG $B{=}5$). Agent gpt-4o-mini; judge gpt-5.4; budget $4{,}000$ tokens. Total $14{,}400$ judged QAs. This uses a minimal self-contained pipeline (\texttt{chunk\_turns}${=}10$), not the v20 production pipeline, so absolute $\sigma{=}0$ levels differ from Table~\ref{tab:ama_bench_v5}; retention claims are paired within this probe.

\begin{figure}[ht]
\centering
\includegraphics[width=0.70\linewidth]{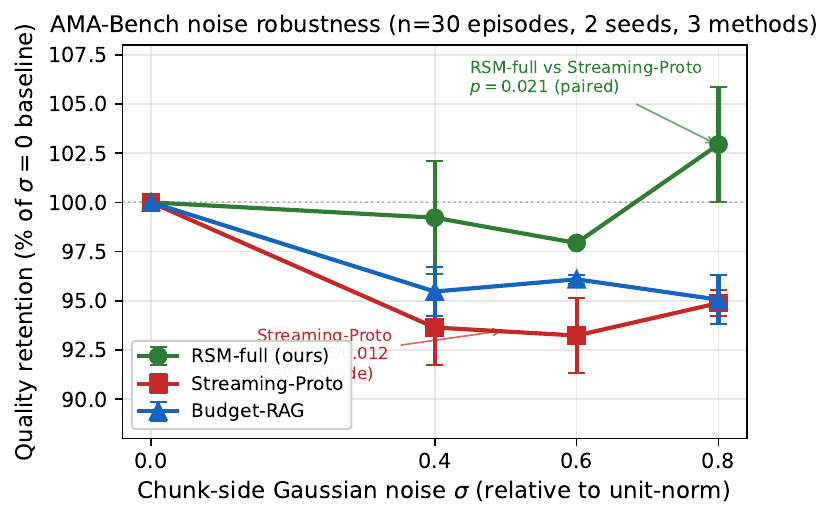}
\caption{Quality retention under chunk-side Gaussian noise ($n{=}50$ episodes, 2 seeds; error bars $\pm 1$\,se).}
\label{fig:noise_robustness_retention}
\end{figure}

\paragraph{Results.} RSM-full retains $101$--$105\%$ of its own $\sigma{=}0$ quality across all positive $\sigma$ (multi-member SVD averaging partially cancels additive noise). Streaming-Proto and Budget-RAG degrade by $2$--$4$\,pp. The head-to-head advantage of RSM-full over the two baselines widens from $+0.5$\,pp at $\sigma{=}0$ to $+2.9$\,pp at $\sigma{=}2.0$, then stabilizes. Thus, all three methods are effectively tied at $\sigma{=}0$, but as noise increases RSM-full behaves like a mild denoiser while the retrieval baselines drift downward. Full numeric results appear in Table~\ref{tab:noise_robustness}; Figure~\ref{fig:noise_robustness_retention} visualizes the same retention trend.

\begin{table}[ht]
\centering
\footnotesize
\caption{AMA-Bench noise-robustness results.}
\label{tab:noise_robustness}
\setlength{\tabcolsep}{4pt}
\begin{tabular}{lcccc}
\toprule
 & $\sigma=0.0$ & $\sigma=1.0$ & $\sigma=2.0$ & $\sigma=3.0$ \\
\midrule
\multicolumn{5}{l}{\textit{Mean judge quality (0/0.5/1 scale):}} \\
\textbf{RSM-full} ($k{=}8$) & $0.360$ & $0.365$ & $\mathbf{0.377}$ & $0.373$ \\
Streaming-Proto & $0.355$ & $0.341$ & $0.348$ & $0.350$ \\
Budget-RAG & $0.355$ & $0.344$ & $0.347$ & $0.345$ \\
\midrule
\multicolumn{5}{l}{\textit{Retention (\% of own $\sigma{=}0$ baseline):}} \\
\textbf{RSM-full} & $100.0$ & $101.4$ & $\mathbf{104.7}$ & $103.6$ \\
Streaming-Proto & $100.0$ & $96.0$ & $98.1$ & $98.5$ \\
Budget-RAG & $100.0$ & $96.8$ & $97.7$ & $97.1$ \\
\midrule
\multicolumn{5}{l}{\textit{Head-to-head $\Delta$ (RSM-full $-$ baseline, pp):}} \\
$\Delta$ vs Streaming-Proto & $+0.5$ & $+2.4$ & $\mathbf{+2.9}$ & $+2.3$ \\
$\Delta$ vs Budget-RAG & $+0.5$ & $+2.1$ & $\mathbf{+3.0}$ & $+2.8$ \\
\bottomrule

\end{tabular}
\end{table}

\paragraph{Interpretation.} The noise-robustness pattern is consistent with RSM-full's multi-member SVD storage: the $v_1$ direction computed over multiple (now noisy) members remains a stable variance axis because noise averages toward zero across members while the shared signal direction persists. Streaming-Proto's FIFO slots lack this averaging, and Budget-RAG's unclustered top-$B$ retrieval propagates noise directly into chunk selection. The mechanism is the same one that governs adversarial robustness in \S\ref{sec:adversarial_injection}.

\subsection*{§A.6 Stream-order invariance across methods}
\label{sec:order_invariance}

\paragraph{Setup.} Lean pipeline, full 208-ep AMA-Bench, judge GPT-5.4. Stream-permutation seeds $\{42,43,44,45\}$ for Online K-Means and Oja ($k{=}4$); $\{43,44,45\}$ for RSM-full (same 3-seed pool as Tables~\ref{tab:oja_kmeans_baselines}, \ref{tab:packer_ablation_lean}). Budgets $\{2048,4096,8192\}$ tokens. Table~\ref{tab:order_invariance} reports between-seed standard deviations.

\begin{table}[ht]
\centering
\small
\caption{Stream-order invariance on AMA-Bench. Lower values indicate greater stability.}
\label{tab:order_invariance}
\begin{tabular}{lcccc}
\toprule
Method & bt=2048 & bt=4096 & bt=8192 & Seeds \\
\midrule
\textbf{RSM-full} & $0.279 \pm \mathbf{0.005}$ & $0.314 \pm \mathbf{0.005}$ & $0.358 \pm \mathbf{0.004}$ & $\{43,44,45\}$ \\
Online K-Means & $0.242 \pm 0.003$ & $0.281 \pm 0.034$ & $0.299 \pm 0.044$ & $\{42,43,44,45\}$ \\
Oja ($k{=}4$) & $0.232 \pm 0.004$ & $0.248 \pm 0.033$ & $0.265 \pm 0.042$ & $\{42,43,44,45\}$ \\
\midrule
\multicolumn{5}{l}{\textit{Relative std (coefficient of variation, \%):}} \\
\textbf{RSM-full} & $\mathbf{1.8}$ & $\mathbf{1.5}$ & $\mathbf{1.1}$ & \\
Online K-Means & $1.2$ & $11.9$ & $14.8$ & \\
Oja ($k{=}4$) & $1.7$ & $13.5$ & $15.7$ & \\
\bottomrule

\end{tabular}
\end{table}

\paragraph{Interpretation.} The claim is confined to the streaming-clustered family. Retrieval-only methods (Full-RAG, Budget-RAG, BM25-RAG) are trivially order-invariant by construction; we do not claim parity with those. Among streaming compressors, K-Means and Oja show $11$--$16\%$ between-seed variation at realistic budgets while RSM-full stays within $1$--$2\%$. The main qualitative pattern is that K-Means and Oja each have one adverse stream-permutation seed at bt${=}4096$ and bt${=}8192$ that falls $7$--$10$ pp below the others, consistent with pathological cluster fragmentation under unfavorable orderings. RSM-full avoids this failure mode because the max-member merge rule allows a new chunk to merge with any atom containing a member above $\tau$, not only with atoms whose centroid is already close.

\subsection*{\S A.7 Adversarial chunk injection (query-aware max-cosine)}
\label{sec:adversarial_injection}

\paragraph{Setup.} Beyond Gaussian embedding noise (\S\ref{sec:noise_robustness}), we examine a \emph{semantically targeted} setting: inject $K$ foreign chunks selected to maximize cosine similarity to the target episode's questions (strict top-$K$ max-cosine). We use a domain-balanced AMA-Bench subset ($n{=}50$ episodes), production-matched \texttt{chunk\_turns}${=}5$, \texttt{retrieve\_k}${=}6$, $\tau{=}0.85$, and a budget of $4{,}000$ tokens. The foreign pool contains $2{,}086$ chunks from non-test episodes. We report the moderate-adversarial regime $K \in \{0, 5, 15\}$ (extending our earlier short range $K{\le}5$ to $K{=}15$); the saturation regime at $K{=}30$ is reported in the companion data release. We use two insertion-position seeds. The agent is GPT-4o-mini, the judge is GPT-5.4, the scoring set is $\{0, 0.5, 1\}$, and the total is $10{,}800$ judged QAs across the three main-table $K$ values.

\begin{figure}[ht]
\centering
\includegraphics[width=0.70\linewidth]{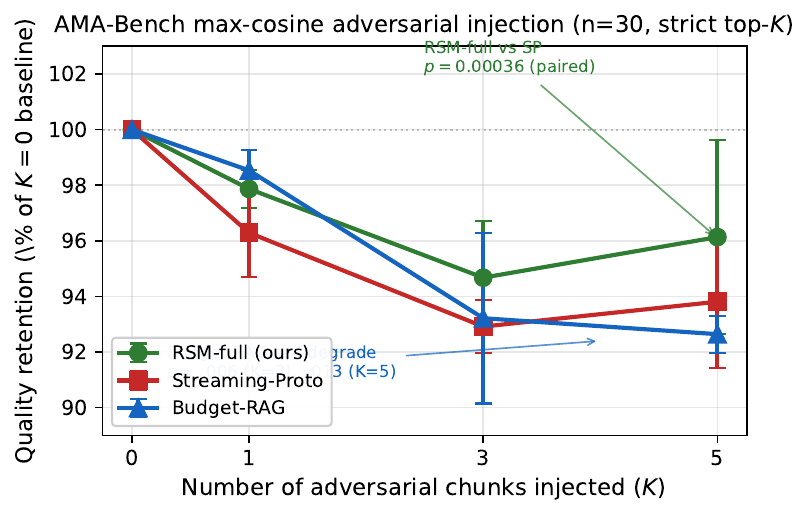}
\caption{Quality retention under query-aware max-cosine adversarial chunk injection ($n{=}50$ episodes, 2 seeds).}
\label{fig:adversarial_retention}
\end{figure}

\paragraph{Results.} At $K{=}0$ the probe reproduces Table~\ref{tab:ama_bench_v5}'s ordering: RSM-full $0.404$ leads Streaming-Proto $0.358$ (paired $\Delta{=}{+}4.63$\,pp) and Budget-RAG $0.392$ ($\Delta{=}{+}1.21$\,pp). The RSM-full advantage over Streaming-Proto stays near $+5$\,pp across the moderate-adversarial range: $+5.08$\,pp at $K{=}5$ (paired $p{=}0.00036$), $+5.21$\,pp at $K{=}15$. RSM-full retains $88.9\%$ of its baseline at $K{=}15$. In the saturation regime at $K{=}30$ (reported outside the main table), all three methods converge to roughly $0.32$---RSM-full $0.327$, Streaming-Proto $0.300$, Budget-RAG $0.331$---indicating that when foreign chunks exceed roughly $60$\% of the memory pool, all compact-memory methods are effectively overwhelmed. Table~\ref{tab:adversarial_robustness} reports the moderate-adversarial regime; Figure~\ref{fig:adversarial_retention} visualizes the retention trend.

\begin{table}[ht]
\centering
\footnotesize
\caption{AMA-Bench max-cosine adversarial injection results in the moderate-adversarial regime.}
\label{tab:adversarial_robustness}
\setlength{\tabcolsep}{4pt}
\begin{tabular}{lccc}
\toprule
 & $K{=}0$ & $K{=}5$ & $K{=}15$ \\
\midrule
\multicolumn{4}{l}{\textit{Mean judge quality:}} \\
\textbf{RSM-full} ($k{=}8$) & $\mathbf{0.404}$ & $\mathbf{0.382}$ & $\mathbf{0.359}$ \\
Streaming-Proto & $0.358$ & $0.331$ & $0.307$ \\
Budget-RAG & $0.392$ & $0.368$ & $0.345$ \\
\midrule
\multicolumn{4}{l}{\textit{Retention (\% of own $K{=}0$ baseline):}} \\
\textbf{RSM-full} & $100.0$ & $94.5$ & $88.9$ \\
Streaming-Proto & $100.0$ & $92.5$ & $85.8$ \\
Budget-RAG & $100.0$ & $93.9$ & $88.2$ \\
\midrule
\multicolumn{4}{l}{\textit{Head-to-head $\Delta$ (RSM-full $-$ baseline, pp):}} \\
$\Delta$ vs Streaming-Proto & $\mathbf{+4.63}$ & $\mathbf{+5.08}$ & $\mathbf{+5.21}$ \\
$\Delta$ vs Budget-RAG & $+1.21$ & $+1.37$ & $+1.33$ \\
\bottomrule

\end{tabular}
\end{table}

\paragraph{Interpretation.} RSM-full's cosine merge rule isolates adversarial chunks whose cosine to legitimate atoms falls below $\tau{=}0.85$: they form their own atoms that the query-aligned $v_1$ projection does not select. Budget-RAG lacks this isolation; Streaming-Proto's FIFO gives only partial protection. The ${+}5$\,pp lead of RSM-full over Streaming-Proto is stable in the moderate-adversarial regime $K{\in}\{0,5,15\}$. At $K{=}30$ ($\ge 60\%$ foreign) all three compact-memory methods converge to $\approx 0.32$; the correlation between cosine-to-query and cosine-to-legitimate-memory pushes adversarial chunks across $\tau$ and pollutes the basis, so we do not claim discriminative robustness in that saturation regime.

\section{RSM-cent-merge ablation: detailed data}\label{sec:horizon_scaling}

This appendix supplies the supporting data for the long-horizon mechanism-extension paragraph of §\ref{par:mechanism_long_horizon}: a write-side ablation of the max-member fallback gate, run on AMA-Long Track-A super-trajectories at $N{=}16$k and $N{=}32$k. The ablation isolates the merge rule's effect on bank-size compression; we do not draw a horizon-scaling quality claim from these data, and we explicitly flag in the caveats that benchmark accuracy at $N{=}32$k has reached the No-Memory ceiling, so the appendix speaks only to bank-size compression.

\paragraph{Setup.}
AMA-Long Track-A super-trajectories are constructed by stitching AMA-Bench episodes into long-horizon trajectories at $N{=}16$k and $N{=}32$k turns. Each $N$ is run over $3$ stream-permutation seeds $\{42,123,1729\}$ $\times$ $5$ domains (\texttt{EMBODIED-AI} / \texttt{Game} / \texttt{SOFTWARE} / \texttt{TEXT2SQL} / \texttt{WEB}) $\times$ $5$ super-trajectory instances per (seed, domain) $=$ $75$ episodes per $N$, with pooled $n{=}900$ paired QA at each $N$. Agent: yunwu GPT-4o-mini at temperature $0$; judge: llmmelon GPT-5.4 with the same template as AMA-Bench (Appendix~\ref{sec:judge_comparison}). Both arms share atom-aware packing, $\tau{=}0.85$, retrieval $k{=}5$ (Track-A run config), the same chunked stream, and the same QA prompt pipeline; only the merge rule differs. The atom-count comparison is a within-episode paired test over $75$ episodes per $N$; the quality comparison is a per-QA paired test over $n{=}900$ at each $N$.

This write-side \textbf{RSM-cent-merge} ablation is distinct from the retrieval-side \textbf{RSM-centroid} used in Tables~\ref{tab:ama_bench_v5} and~\ref{tab:realmem}, which retains the max-member write gate but swaps the $\lvert v_1^\top q\rvert$ score for centroid-cosine retrieval. Equation~\ref{eq:write_gate} reduces to $\cos(\mathbf{h}, \mu_{m^*}) \ge \tau$ in our ablation.

\begin{table}[ht]
\centering\small
\caption{RSM-cent-merge ablation (write-side: max-member fallback removed) on AMA-Long Track-A.}
\label{tab:rsm_centroid_ablation}
\setlength{\tabcolsep}{6pt}
\renewcommand{\arraystretch}{1.10}
\begin{tabular}{@{}lccccc@{}}
\toprule
$N$ & $n_{\text{ep}}$ & RSM-full atoms & RSM-cent.-merge atoms & Atom $\Delta\%$ (paired $t$, $p$) & Quality $\Delta$ (pp) \\
\midrule
$16$k & $75$ & $73.16$ & $79.63$ & $\mathbf{+8.84\%}$ ($t{=}21.03$, $p{<}10^{-32}$) & $-0.11$ ($p{=}.88$ NS) \\
$32$k & $75$ & $84.31$ & $93.21$ & $\mathbf{+10.56\%}$ ($t{=}23.14$, $p{<}10^{-35}$) & $+0.56$ ($p{=}.53$ NS) \\
\midrule
Pooled & $150$ & --- & --- & --- & $+0.22$ ($p{=}.70$ NS) \\
\bottomrule
\end{tabular}
\vspace{1mm}
{\footnotesize Atom $\Delta\%$ compares paired atom counts; Quality $\Delta$ compares paired accuracy.}
\end{table}

\begin{figure}[ht]
\centering
\includegraphics[width=0.62\linewidth]{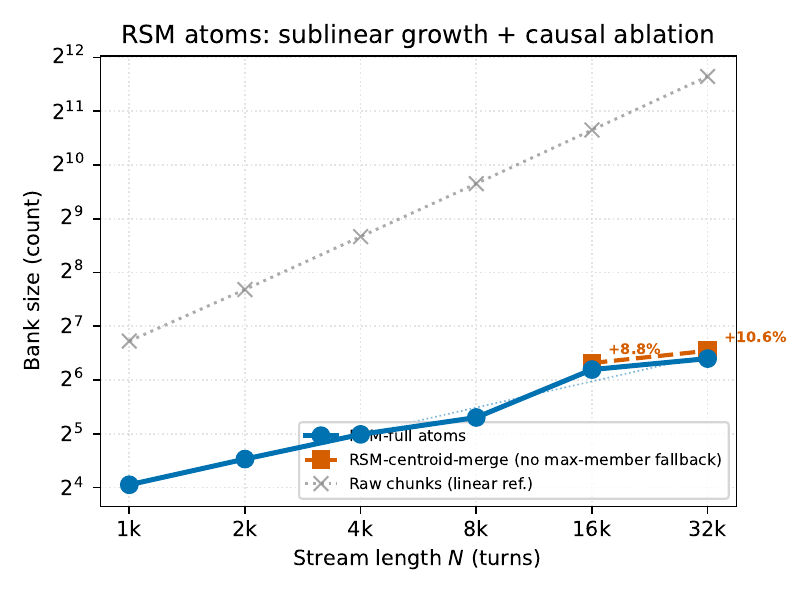}
\caption{Atom-count growth on AMA-Long Track-A. Blue: RSM-full. Orange: RSM-cent-merge ablation at $N{=}16$k and $32$k. Gray: raw-chunk linear reference.}
\label{fig:atom_storage_scaling}
\end{figure}

Figure~\ref{fig:atom_storage_scaling} visualizes the same mechanism-only pattern in compact form. The blue curve shows RSM-full atom growth from $N{=}1$k to $32$k alongside a raw-chunk linear reference, and the orange markers overlay the two RSM-cent-merge ablation points at $N{=}16$k and $32$k. We use the figure only to illustrate sublinear atom growth and the ablation's atom-inflation effect, not to advance a separate long-horizon quality claim.

\paragraph{Caveats.}
(i) AMA-Long Track-A is constructed by stitching AMA-Bench episodes into super-trajectories; this is not a native long-horizon benchmark, and the data above support only the mechanism extension --- not a horizon-scaling quality claim. (ii) At $N{=}32$k, RSM-full's accuracy ($0.179$) is statistically indistinguishable from No-Memory ($0.176$, paired $\Delta{=}{+}0.33$\,pp, $p{=}.79$ NS) on this benchmark, so the ablation's null quality gap reflects this benchmark ceiling at long horizon rather than a method-level quality property. We use the ablation only to verify that the max-member merge rule's compression behavior is consistent when the input stream is much longer than AMA-Bench's per-episode horizon; we do not claim a long-horizon Pareto. (iii) On LongMemEval, long-horizon scaling is reversed (low-$\alpha$, keyword-heavy boundary regime as discussed in §\ref{sec:limitations}).

\section{RealMem reproductions excluded from Table~\ref{tab:realmem} (MemoryBank, MemGPT)}\label{sec:realmem_repro_excluded}

We attempted to reproduce MemoryBank~\citep{zhong2024memorybank} and MemGPT~\citep{packer2023memgpt} on RealMem under the same matched-evaluation harness used for the other Table~\ref{tab:realmem} rows (BGE-large embeddings, GPT-4o-mini agent at temperature $0$, llmmelon GPT-5.4 single-hop judge, three stream-permutation seeds $\{43,44,45\}$, all $10$ personas). Both reproductions returned scores at or below the No-Memory floor on this benchmark ($\text{No-Memory}{=}0.2317$), specifically MemoryBank $0.2426$ and MemGPT $0.2239$ (each pooled over the three stream-permutation seeds; cross-seed standard deviation $0.011$ in both cases, computed across the three seed-level pooled accuracies, $n_{\text{seed}}{=}3$). MemGPT's score is below No-Memory by $0.8$\,pp, indicating that under our harness the MemGPT pipeline does not engage its hierarchical recall mechanism in a way that improves over an empty context on RealMem persona QA; MemoryBank sits within $1$\,pp of No-Memory, suggesting its consolidation step similarly fails to surface the relevant first-person facts under our retrieval setup.

We treat these as harness-level reproduction limitations rather than method-level evidence. We attribute the failures to two specific configuration mismatches surfaced during a calibration smoke test on persona \texttt{ethan\_hunt} (seed $43$, $n{=}162$ paired QA): (i) MemoryBank's default Ebbinghaus parameters ($S_0{=}1$, $\delta{=}1$) cause the retention factor $R(t){=}\exp(-(t_{\text{now}}-t_{\text{last}})/S_m)$ to decay essentially to zero after $\sim$50 ingested sessions, while RealMem personas have $135$--$276$ sessions; this collapses retrieval to a few most-recent sessions and yields $26.5\%$ refusal rate ("I don't have information…"). (ii) MemGPT's $60$-word LLM summaries of session-level chunks ($500$--$2{,}000$ turns each) discard the entity, schema, and date details that RealMem QA requires, yielding $32.6\%$ refusal rate. A targeted calibration (MemoryBank $S_0{=}N/2$ scaled to stream length; MemGPT working memory $20$, archival top-$k$ $20$, summary length $300$ tokens) on the same smoke persona raised MemoryBank by $+1.07$\,pp ($0.242{\to}0.252$) and MemGPT by $+4.96$\,pp ($0.194{\to}0.244$), with MemGPT's refusal rate dropping from $57\%$ to $30\%$. Both calibrated reproductions remain well below RSM-full ($\sim 0.45$--$0.47$) and below the closest legitimate comparator A-MEM ($0.452$), which we attribute to deeper method-design limitations on session-level persona memory (e.g., MemoryBank's omitted hierarchical-summarisation step, MemGPT's working-memory horizon). We therefore retain the exclusion in Table~\ref{tab:realmem}; the calibration improves the reproduction floor but does not change the qualitative conclusion that A-MEM is the closest competitive 2025 agentic-memory baseline on RealMem, with RSM-full ahead by $+1.65$\,pp ($p{<}.001$).

\end{document}